%% file: acl_latex.tex
\pdfoutput=1

\documentclass[11pt]{article}

\usepackage[]{acl}

\usepackage{times}
\usepackage{latexsym}

\usepackage[T1]{fontenc}

\usepackage[utf8]{inputenc}

\usepackage{microtype}

\usepackage{inconsolata}

\usepackage{graphicx}

\usepackage{xcolor,colortbl}
\usepackage{nicematrix} 
\usepackage{multirow}
\usepackage{amssymb,mathtools}
\usepackage{makecell,booktabs}
\usepackage{cleveref}
\crefformat{section}{\S#2#1#3}
\crefformat{subsection}{\S#2#1#3}
\title{Why Do Some Inputs Break Low-Bit LLM Quantization?}

\author{
\textbf{Ting-Yun Chang}
\quad\quad \textbf{Muru Zhang}
\quad\quad \textbf{Jesse Thomason}
\quad\quad \textbf{Robin Jia} \\
University of Southern California, Los Angeles, CA, USA \\
\texttt{\{tingyun, muruzhan, jessetho, robinjia\}@usc.edu}
}

\begin{document}
\maketitle

\input{dfn.tex}
\definecolor{bgyellow}{RGB}{255, 251, 240}
\definecolor{bggreen}{RGB}{240, 255, 251}
\definecolor{bgblue}{RGB}{240, 244, 255}
\definecolor{bgpurple}{RGB}{251, 240, 255}

\begin{abstract}
\input{00_abstract}
\end{abstract}
\input{01_intro}
\input{02_related_work}
\input{03_setups}
\input{04_different_methods}
\input{05_destiny}
\input{06_where}
\input{07_data_analysis}
\input{08_conclusion}
\input{09_limitations}
\input{acknowledgements.tex}
\bibliography{custom}

\clearpage
\appendix
\input{10_appendix}
\end{document}

%% file: dfn.tex
\newcommand{\NLL}{\text{NLL}}
\newcommand{\Err}{e}
\newcommand{\LLM}{\theta}
\newcommand{\QLLM}{\tilde{\LLM}}
\newcommand{\EE}{p^{(l)}}
\newcommand{\EENLL}{\text{NLL}^{(l)}}
\newcommand{\W}{W^{(l)}}

\newcommand{\QW}{\tilde{W}^{(l)}}
\newcommand{\Gate}{W_{\text{gate}}}
\newcommand{\Up}{W_{\text{up}}}
\newcommand{\Down}{W_{\text{down}}}
\newcommand{\HGate}{h_{\text{gate}}}
\newcommand{\HUp}{h_{\text{up}}}
\newcommand{\HDown}{h_{\text{down}}}
\newcommand{\HAttn}{h_{\text{attn}}}

\newcommand{\Out}{o^{(l)}}
\newcommand{\QOut}{\tilde{o}^{(l)}}
\newcommand{\Ds}{\mathcal{D}}
\newcommand{\EDs}{\mathcal{E}}
\newcommand{\corr}{\rho}
\newcommand{\Ex}{x}
\newcommand{\Hidden}{x}
\newcommand{\MLP}{z^{(l)}}
\newcommand{\PrevMLP}{z^{(l-1)}}
\newcommand{\Simple}{r^{(l)}}
\newcommand{\Simplet}{r_t^{(l)}}
\newcommand{\PostLN}{h^{(l)}}
\newcommand{\RR}{\mathbb{R}}
\newcommand{\norm}{\operatorname{norm}}
\newcommand{\Norm}{\mathbb{S}}
\newcommand{\Exnorm}{\norm \left(\Ex; \Simple \right)}
\newcommand{\Expostnorm}{\norm \left(\Ex; \PostLN \right)}
\newcommand{\NoiseExpost}{\norm (\Ex; \tilde{h}^{(l)} )}
\newcommand{\LargeD}{\mathcal{D}_{\text{large}}}
\newcommand{\Ctrl}{\mathcal{D}_{\text{ctrl}}}
\newcommand{\nll}{\operatorname{nll}}

%% file: 00_abstract.tex
Low-bit weight-only quantization significantly reduces the memory footprint of large language models (LLMs), but disproportionately affects certain examples.
We analyze diverse 3--4 bit methods on LLMs ranging from 7B--70B in size and 
find that the quantization errors of 50 pairs of methods are strongly correlated (avg. $\corr = 0.82$) on FineWeb examples.
Moreover, the residual stream magnitudes of full-precision models are indicative of future quantization errors.
We further establish a hypothesis that relates the residual stream magnitudes to error amplification and accumulation over layers.
Using LLM localization techniques, early exiting, and activation patching, we show that examples with large errors rely on precise residual activations in the later layers, and that the outputs of MLP gates play a crucial role in maintaining the perplexity.
Our work reveals why certain examples result in large quantization errors and which model components are most critical for performance preservation.

%% file: 01_intro.tex
\section{Introduction}
\label{sec:intro}
Many large language models (LLMs; \citealp{llama3modelcard,team2024gemma,olmo20242olmo2furious,qwen3}) use 16-bit precision to represent each parameter by default, requiring substantial GPU memory for inference.
Quantization reduces memory usage by reducing the bit-precision of model weights or activations.
Low-bit weight-only quantization \cite{frantar2023optq,pmlr-v202-sheng23a,park2024lutgemm,lee2024owq}, which quantizes model weights to $\le 4$ bits and keeps the activations in 16 bits, greatly reduces the memory footprint while maintaining reasonable performance across various benchmarks \cite{kurtic2024give}.

However, \citet{dettmers2023case} find that at 3-bits, LLMs start to show noticeable accuracy degradation.
\citet{kumarscaling} show that LLMs trained on more data become harder to quantize.
Prior efforts focus on \emph{outliers}, the abnormally large activations or weights that cause large errors in LLM quantization \cite{dettmers2022gptint}.
Quantization methods that combat outliers lead to overall better performance \cite{xiao2023smoothquant,chee2023quip,ashkboos2024quarot,luo2025fast}; however, we observe that some examples still suffer from large quantization errors under various methods.
Why does quantization disproportionately affect certain examples?
What factors are indicative of quantization errors?
Which parts of the LLMs lead to large errors?
We study these underexplored questions across diverse 3- and 4-bit weight-only quantization methods.

First, we observe that the quantization errors of various pairs of methods are highly correlated (avg. $\corr = 0.82$) and that certain examples yield large errors across methods, where we define errors as the increase in negative log-likelihoods on sampled pretraining documents after quantization.
A possible hypothesis is that the errors are largely predetermined by the full-precision model.
Since \citet{awq2024} show that activation outliers amplify the errors in quantized weights, we examine whether the activation magnitudes of the full-precision models are indicative of the future quantization errors. 

Surprisingly, we find strong \emph{negative} correlations between quantization errors and the magnitudes of the residual stream.
For instance, the correlations computed on the residuals across the last 30 layers of Llama3-70B often exceed $-0.6$ and reach $-0.8$ in the final layer.
The strong correlations support our hypothesis that the full-precision model largely decides the future quantization errors.
Meanwhile, it is unexpected that smaller residual activations are associated with large quantization errors.
Our further analysis reveals that the RMS LayerNorm \cite{zhang2019root} tends to \emph{reverse} the relative activation magnitudes.
These findings altogether explain why certain examples have large quantization errors across methods: examples that inherently have smaller residual magnitudes will have larger post-RMSNorm magnitudes over multiple layers, which continually amplify the errors in the quantized weights and ultimately result in large quantization errors in outputs.

We further apply LLM localization techniques to study where quantization errors arise from.
First, we use early exiting \cite{nostalgebraist} to project the residual hidden states across layers to the output probabilities.
Early exiting reveals that examples with large quantization errors rely heavily on later layers to adjust the model output distributions.
However, restoring the weights of later layers to 16 bits only marginally reduces perplexity, implying that the problem lies with the accumulated errors in activations over layers.
Next, we adapt cross-model activation patching \cite{prakash2023fine}, which restores specific activations of the quantized model to the clean counterparts from the full-precision model.
We find that restoring the outputs of the MLP gate projection \cite{shazeer2020glu} can substantially reduce the perplexity.

Finally, we investigate what kinds of data tend 
to have large quantization errors.
We classify the large-error examples and find that they cover diverse topics.
Because \citet{ahia-etal-2021-low-resource,marchisio2024-quantization} show that model compression has a disparate impact on long-tail examples, we also study the relationship between quantization errors and long-tail examples.
Surprisingly, our results on both FineWeb \cite{penedo2024the} and PopQA \cite{mallen-etal-2023-trust} datasets suggest that quantization has a milder impact on long-tail examples and that well-represented examples are not immune to large degradation in log-likelihood and accuracy.

Our work centers on why, where, and what leads to large quantization errors, offering systematic analyses across diverse methods.
We provide a mechanistic explanation of why certain examples suffer from large quantization errors and identify key model components that are critical to maintaining the perplexity.
Our study could motivate future methods to reduce errors from MLP gate projections or to introduce LayerNorm scaling factors \cite{wei2023outlier,awq2024} that consider the reversal effects in RMSNorm.\footnote{Code: \url{https://github.com/terarachang/QError}}

%% file: 02_related_work.tex
\section{Background}
\label{sec:background}


\label{sec:q_methods}
This section introduces the quantization methods in our study, covering diverse, widely-used approaches for low-bit weight-only quantization.

\paragraph{GPTQ.} \citet{frantar2023optq} propose GPTQ, an approximate second-order method that quantizes one
weight at a time while updating not-yet-quantized weights to compensate for the quantization loss.
GPTQ minimizes the loss on a small calibration set of C4 \cite{raffel2020exploring} samples.
We adopt common GPTQ settings, using activation sorting and a Hessian damping coefficient of 0.01 when not otherwise specified.

\paragraph{AWQ.} \citet{awq2024} observe that the activation outliers magnify the quantization errors in the corresponding weight dimensions. They propose AWQ, which smoothes the activations by introducing a scaling factor corresponding to each input dimension of weights.
AWQ grid-searches the scaling factors based on the activation magnitudes.

\paragraph{NormalFloat.} \citet{dettmers2023qlora} propose NormalFloat (NF), a 4-bit data type that is optimal for normally distributed weights based on information theory.
NF employs non-uniform quantization with Gaussian quantiles and requires a lookup table.
\citet{guo-etal-2024-fast} accelerate the lookup table operation and extend NF to lower bit precision.

\paragraph{EfficientQAT.} \citet{efficientqat} propose EfficientQAT (EQAT), a lightweight quantization-aware fine-tuning \cite{liu-etal-2024-llm} method.
EQAT applies straight-through estimator \cite{bengio2013estimating} for backpropagation.

\paragraph{LayerNorm in Quantization.} \citet{kovaleva-etal-2021-bert,wei2022outlier} find that outliers in the LayerNorm parameters of BERT \cite{bert} cause difficulties in model compression.
Given the importance of LayerNorm, all the quantization methods we discuss above leave LayerNorm unquantized.
Specifically, LLMs in Llama 
\cite{touvron2023llama}, Mistral \cite{jiang2023mistral}, and Qwen \cite{qwen} families use RMSNorm \cite{zhang2019root}:
\begin{align}
    \text{RMSNorm}(z) &= \frac{z}{\text{RMS}(z)}\odot \gamma, \label{eq:rms} \\
    \text{RMS}(z) &= \sqrt{\frac{1}{d}\sum_{i=1}^{d} z_i^2},
\end{align}
where $z$ is the $d$-dimensional input, $\odot$ denotes element-wise multiplication, and $\gamma \in \RR^d$ is the unqauntized weights for affine transformation.

%% file: 03_setups.tex
\section{Problem Setups}
\label{sec:setups}
\subsection{Datasets}
\label{sec:fineweb}
\paragraph{FineWeb.} 
We create a dataset $\Ds$ of 10k examples sampled from the FineWeb corpus \cite{penedo2024the}, consisting of deduplicated English web documents.
We evaluate the negative log-likelihood (NLL) of different models on $\Ds$.
Formally, given a model $\LLM$ and an example $\Ex = [\Ex_0, \Ex_1, \dotsc, \Ex_T]$ of $T$ tokens, prepended with a start of sentence token $\Ex_0$, we define the NLL of $\Ex$ as:
\begin{equation}
    \NLL(\Ex; \LLM) = \frac{1}{T}\sum_{t=1}^{T} - \log p_\LLM (\Ex_t | \Ex_{ < t}), \label{eq: lm_loss}
\end{equation}
where $p_\LLM$ is the probability distribution of the model $\LLM$.
We only sample from long documents and truncate all the sampled documents to $512$ tokens.

\paragraph{Quantization Errors.} NLL is a common loss term for LLM pretraining, and prior quantization research uses perplexity (exponentiated average NLL) as a standard metric.
Therefore, we define the quantization error on a FineWeb example $x$ as the increase in NLL.
Formally, given the full-precision LLM $\LLM$ and its quantized counterpart $\QLLM$, we define the quantization error of $\QLLM$ on $x$ as:
\begin{equation}
    \Err(\Ex; \QLLM) = \NLL(\Ex; \QLLM) - \NLL(\Ex; \LLM) \label{eq: error}
\end{equation}

\paragraph{Large-Error Set and Control Set.} For fine-grained analyses, we create a large-error set $\LargeD \subset \Ds$, consisting of the top-$10\%$ (1k) examples by quantization errors and a control set $\Ctrl$ of 1k examples sampled from the bottom-$50\%$.\footnote{See \cref{sec:app_dlarge} for detailed dataset construction.}


\paragraph{PopQA.} We use PopQA \cite{penedo2024the} to study how quantization affects LLMs' factual knowledge of entities under different popularity levels.
PopQA contains 14k questions, each supplemented with the entity popularity. 
Following \citet{penedo2024the}, we use 15-shot prompting, greedy decoding, and exact match accuracy.

\subsection{Models and Quantization Methods}
We study four LLMs: Qwen2.5-7B \cite{qwen2.5}, Llama3-8B \cite{llama3modelcard}, Mistral-Nemo-12B,\footnote{\url{https://mistral.ai/news/mistral-nemo}} and Llama3-70B.
We study quantization methods in the 3--4 bit, weight-only quantization regime: GPTQ, AWQ, NF, and EQAT (see \cref{sec:background}).
We follow original implementations and detail the calibration sets, quantization configurations, and perplexity of each method in \cref{sec:app_details}.

%% file: 04_different_methods.tex
\input{latex/Tables/corr_main}
\section{Do Methods Fail on the Same Inputs?}

\label{sec:diff_q}
First, we investigate whether diverse quantization methods cause large errors on the same inputs. 
Recall that we define the example-level quantization error $\Err(\Ex^{(i)}; \QLLM) \in \RR$ in Eq.~\ref{eq: error}.
Given a dataset $\Ds = \{\Ex^{(i)}\}_{i=1}^N$ of $N$ examples, 
we define the errors of the quantized model $\QLLM$ on the dataset $\Ds$ as:
\begin{equation}
    \EDs(\Ds ; \QLLM) = [\Err(\Ex^{(1)};\QLLM), \dotsc, \Err(\Ex^{(N)};\QLLM)] \; \in \RR^N, \nonumber \label{eq:ds_error}
\end{equation}

Let $\QLLM_1$ and $\QLLM_2$ be the quantized models of the same base LLM $\LLM$, produced using different methods.
We compute their quantization errors on the dataset, $\EDs_1 \triangleq \EDs(\Ds; \QLLM_1)$ and $\EDs_2 \triangleq \EDs (\Ds; \QLLM_2)$, respectively, and report the Pearson correlation between the errors, denoted as $\corr (\EDs_1, \EDs_2)$.

\subsection{Strong Correlations Between Methods}
Table~\ref{table:fineweb_corr} shows the correlations between the quantization errors of every two methods. 
We include the results of Qwen2.5-7B and Llama3-8B in Table~\ref{table:app_fineweb_corr} in appendix.
Across four LLMs, we observe high correlations, an average of $\corr (\EDs_1,\EDs_2) = 0.82$ over all 50 pairs of methods, although Qwen2.5-7B has lower correlations (avg. $0.66$) compared to other LLMs.
The overall strong correlation indicates that diverse methods have a similar effect on LLMs' likelihoods.
We further find that the top-$10\%$ large-error examples of different methods on the same LLMs are highly overlapped, with an average Jaccard similarity of $0.51$, far above the~$\sim 0.05$ expected values by chance.

Since the configurations of the GPTQ algorithm can greatly affect the quantization outcomes~\cite{frantar2023optq,chen2025geometry}, we further examine correlations in quantization errors across different GPTQ variants.
We focus on two factors: (1) the damping coefficient that improves the numerical stability of the inverse Hessian, varied over the range $[0.005, 0.01, 0.02]$, and (2) whether activation-based sorting is applied to change the weight quantization order.
We find that the 3-bit Qwen2.5-7B model variants produced by different GPTQ configurations show a strong average correlation ($0.93$) in quantization errors on FineWeb (see Table~\ref{table:app_gptq} in the appendix).

%% file: latex/Tables/corr_main.tex
\begin{table}[!b]
\begin{center}
\begin{small}
\begin{NiceTabular}{llc}
    \CodeBefore
    \rectanglecolor{bggreen}{1-1}{26-1}
    \rectanglecolor{bgyellow}{1-2}{26-2}
    \rectanglecolor{bgblue}{1-3}{26-3}
    \Body
\toprule
\textbf{Model} & \textbf{(Method1, Method2)} & \boldmath{$\corr (\EDs_1, \EDs_2)$} \\
    
\midrule
\textbf{Mistral-12B}
& ( AWQ4 , NF4   ) & 0.78 \\
& ( AWQ4 , GPTQ4 ) & 0.86 \\
& ( AWQ4 , NF3   ) & 0.80 \\
& ( AWQ4 , GPTQ3 ) & 0.83 \\
& ( AWQ3 , NF3   ) & 0.90 \\
& ( AWQ3 , GPTQ3 ) & 0.95 \\
& ( NF4  , GPTQ4 ) & 0.82 \\
& ( NF4  , AWQ3  ) & 0.81 \\
& ( NF4  , GPTQ3 ) & 0.78 \\
& ( NF3  , GPTQ3 ) & 0.89 \\
& ( GPTQ4, AWQ3  ) & 0.90 \\
& ( GPTQ4, NF3   ) & 0.85 \\
\midrule
\textbf{Llama3-70B}
& ( AWQ4 , NF4   ) & 0.80 \\
& ( AWQ4 , GPTQ4 ) & 0.96 \\
& ( AWQ4 , GPTQ3 ) & 0.86 \\
& ( AWQ4 , EQAT3 ) & 0.81 \\
& ( AWQ3 , EQAT3 ) & 0.85 \\
& ( NF4  , GPTQ4 ) & 0.81 \\
& ( NF4  , GPTQ3 ) & 0.80 \\
& ( NF4  , AWQ3  ) & 0.81 \\
& ( NF4  , EQAT3 ) & 0.75 \\
& ( GPTQ4, AWQ3  ) & 0.94 \\
& ( GPTQ4, EQAT3 ) & 0.83 \\
& ( GPTQ3 , AWQ3 ) & 0.96 \\
& ( GPTQ3, EQAT3 ) & 0.83 \\

\bottomrule
\end{NiceTabular}
\end{small}
\end{center}
\caption{The quantization errors of different methods are strongly correlated on the FineWeb. The numbers in the method names denote 3- or 4-bit precision.}
\label{table:fineweb_corr}

\end{table}

%% file: 05_destiny.tex
\section{Why Do Examples Have Large Errors?}
\label{sec:destiny}

We further explore why certain examples have large quantization errors across methods.
Inspired by \citet{awq2024}, we first study whether the activation magnitudes from the full-precision models are predictive of the quantization errors.

\subsection{Notations}
Formally, a Transformer layer \cite{vaswani2017attention} in modern LLMs consists of a multi-headed attention (MHA), an MLP, and two Pre-LNs \cite{preln}, LN$_1$ and LN$_2$:
\begin{align}
& \Simple = \PrevMLP + \operatorname{MHA}^{(l)} \left(\operatorname{LN}^{(l)}_1 (\PrevMLP) \right), \label{eq:residual} \\
& \MLP = \Simple + \operatorname{MLP}^{(l)} \left( \operatorname{LN}^{(l)}_2 (\Simple) \right), \label{eq:residual_mlp}
\end{align}
where $l$ is the layer index and $z^{(0)} \in \RR^d$ is the input to the first layer.
We observe that the representations after the residual operations, $\Simple$ and $\MLP$, are both indicative of quantization errors.
As $\Simple$ shows clearer patterns, we name it as residual state and include the results of $\MLP$ in appendix.
Note that all the studied methods quantize the weights in MHA and MLP, but keep the weights $\gamma$ in LN$_1$ and LN$_2$ in 16 bits (see \cref{sec:background}).

Given an example $\Ex$ of $T$ tokens, we quantify the magnitude of its residual states at layer $l$ by averaging the Euclidean norm of each token representation $\Simplet \in \RR^d$ over all $T$ tokens:

\begin{align}
    \Exnorm = \frac{1}{T} \sum_{t=1}^T \|\Simplet\|_2 \label{eq:example_norm}
\end{align}
We name $\Exnorm \in \RR$ as the residual magnitude of example $\Ex$ at layer $l$. 
Given a dataset $\Ds = \{\Ex^{(i)}\}_{i=1}^N$ of $N$ examples, we compute $\Exnorm$ for each example $\Ex^{(i)}$ and denote the residual magnitudes on $\Ds$ as $\Norm (\Ds; \LLM; l ) \in \RR^N$.

Finally, given an full-precision model $\LLM$ with a layer index $l$, a quantized model $\QLLM$, and a dataset $\Ds$ of $N$ examples, we calculate the Pearson correlation between 
the residual magnitudes $\Norm (\Ds; \LLM; l)$ and the quantization errors $\EDs(\Ds ; \QLLM)$, denoted as $\corr \left(\Norm(l), \EDs \right)$.
Note that $\EDs(\Ds; \QLLM)$ measures the final errors of the quantized model and thus does not depend on the layer index $l$.

\input{latex/Tables/uq_err_corr}
\input{latex/Figures/density}
\subsection{Residual Magnitudes From Full-Precision Models Indicates Quantization Errors}
We observe a strong negative correlation between the quantization error $\EDs(\Ds ; \QLLM)$ and the residual magnitudes $\Norm (\Ds; \LLM; l)$ computed on the last few layers of the full-precision model, where $\Ds$ is the FineWeb dataset of 10k examples.
For example, let $l := L$ be the last Transformer layer, we report the correlation $\corr \left(\Norm(L), \EDs \right)$ of different LLMs and quantization methods in Table~\ref{table:uq_err_corr}.
The average correlation over 3-bit methods on Qwen2.5-7B, Llama3-8B, Mistral-12B, and Llama3-70B are $-0.66, -0.76, -0.78$, and $-0.80$.
As shown in \cref{sec:app_corr_res_err}, the correlations become increasingly negative in upper layers.
The strong negative correlations suggest that (1) the full-precision LLMs can indicate an example’s future quantization error, and (2) small residual magnitudes in upper layers are associated with large quantization errors.

To further investigate the cause of large errors, we compare the large-error set $\LargeD$ with the control set $\Ctrl$ (\cref{sec:fineweb}).
We compute the average residual magnitudes over examples in $\LargeD$ and $\Ctrl$, respectively, and then compare their average magnitudes across the upper half layers of Llama3-70B.
In Fig.~\ref{fig:main-density} (a), we find that for both datasets, the average residual magnitudes increase over layers; however, the $\LargeD$ (orange) consistently has smaller magnitudes than $\Ctrl$ (blue).
This implies that having a small residual magnitude in the final layer is the cumulative result of continually having smaller residual magnitudes over the upper layers.
Our observation seems to contradict the previous belief that \emph{large} activations are detrimental to weight-only quantization \cite{awq2024}. 
We investigate the cause of this discrepancy in the next section.

\subsection{From Residuals to Errors}
\paragraph{The reversal effect of RMSNorm.} Why are smaller residual magnitudes in upper layers associated with larger quantization errors?
The RMSNorm applied to the residual states before MLP (Eq.~\ref{eq:residual_mlp}), plays a crucial role.
We observe a ``reversal effect’’, where the activations with smaller magnitudes often end up with larger magnitudes after applying RMSNorm.
Formally, we denote the hidden state after RMSNorm as \mbox{ $\PostLN = \text{RMSNorm}^{(l)}(\Simple)$ }.
Following Eq.~\ref{eq:example_norm}, we can compute the magnitude of $\PostLN$, denoted as $\Expostnorm$.
Fig.~\ref{fig:main-density} (b) \& (c) present the density of $\Exnorm$ and $\Expostnorm$, respectively.
Initially, $\LargeD$ has smaller $\Exnorm$ than $\Ctrl$, but after RMSNorm, it has larger $\Expostnorm$.
We observe the reversal effect in the upper layers of all four LLMs.

\paragraph{Activations amplify errors in weights.} 
Since the hidden state after RMSNorm is the immediate input to the quantized MLP (Eq.~\ref{eq:residual_mlp}), we hypothesize that a larger $\Expostnorm$ leads to amplification of errors in the quantized weights.
Let $\PostLN$ and $\tilde{h}^{(l)}$ be the inputs to the $l$-th layer MLP in the full-precision and quantized models, respectively, and let $\Out$ and $\QOut$ be the corresponding MLP outputs.
We compute the mean squared error between $\Out$ and $\QOut$ to quantify the layer-wise output error, denoted as $\operatorname{MSE}^{(l)}$.
We confirm that the input magnitudes $\Expostnorm$ have positive correlations ($>0.6$) with $\operatorname{MSE}^{(l)}$ in most upper layers.
\input{latex/Tables/notations}
\paragraph{Full Hypothesis.} 
Combining our findings above, we establish a full hypothesis: certain examples inherently have smaller residual magnitudes over layers.
Due to the reversal effect, they tend to have larger activation magnitudes after RMSNorms.
When the model is quantized, large activations will amplify the quantization errors in the weights over multiple layers and eventually result in large quantization errors.
Both the residual states and errors are cumulative over layers, which may explain why the final-layer residual magnitudes $\Norm (\Ds; \LLM; L)$ have the strongest correlation with the final errors $\EDs(\Ds ; \QLLM)$.

\paragraph{Why does $\LargeD$ have smaller residual magnitudes?}
We find that examples in the large-error set have fewer/weaker outliers in the residual states $\Simple$. 
Following \citet{bondarenko2023quantizable,gong2024llmc}, we use the Kurtosis value to quantify the severity of outliers.
We measure the Kurtosis value $\kappa^{(l)}$ of the $l$-th layer residual states $\Simple$, where $\kappa^{(l)} =\mathbb{E} \left[ \left( \frac{\Simple - \mu}{\sigma} \right)^4 \right]$.
A large $\kappa^{(l)}$ indicates more outliers \cite{liu2025spinquant}.
Fig.~\ref{fig:app-kappa} in appendix shows that the Kurtosis values of $\LargeD$ (orange) are smaller than those of  $\Ctrl$ (blue) across the upper layers of LLMs, suggesting that $\LargeD$ has fewer outliers in the residual states.
Because outliers dominate the activation magnitudes, having fewer outliers leads to smaller residual magnitudes.

\paragraph{How does RMSNorm reverse the relative magnitudes?}
We observe that the RMSNorm weights $\gamma \in \RR^d$ suppress many outliers in the residual states by assigning smaller weight values to the corresponding dimensions.
Specifically, we rank the entries in $\gamma$ by their absolute values and find that in the last few layers, the largest outliers\footnote{The ``most-activated’’ dimension in $\Simple$ that has the highest average absolute activation over the FineWeb dataset.} often have the smallest RMSNorm weights across all four LLMs.
We include the statistics and discussion on other outlier dimensions in \cref{sec:app_rms_outlier}.
We summarize the mechanism behind the reversal effect: a residual state $\Simple$ with fewer outliers has a smaller magnitude, but RMSNorm normalizes the magnitude and downweights outliers with $\gamma$, yielding a hidden state $\PostLN$ with a relatively larger magnitude.
Since $\LargeD$ examples have smaller kurtosis values but larger $\Expostnorm$, the large quantization errors are driven by the overall larger activations in $\PostLN$.

%% file: latex/Tables/uq_err_corr.tex
\begin{table}[!b]
\begin{center}

\begin{NiceTabular}{llc}
    \CodeBefore
    \rectanglecolor{bggreen}{1-1}{12-1}
    \rectanglecolor{bgyellow}{1-2}{12-2}
    \rectanglecolor{bgblue}{1-3}{12-3}
    \Body
\toprule
\textbf{Model} & \textbf{Method} & \boldmath{$\corr \left(\Norm(L), \EDs \right)$} \\
\midrule
\textbf{Qwen2.5-7B} 
& AWQ3 & -0.66 \\
& NF3 & -0.58 \\
& GPTQ3 & -0.74 \\
\midrule
\textbf{Llama3-8B}
& AWQ3 & -0.76 \\
& NF3 & -0.76 \\
& EQAT3 & -0.75 \\

\midrule
\textbf{Mistral-12B}
& AWQ3 & -0.80 \\
& NF3 & -0.71 \\
& GPTQ3 & -0.83 \\
\midrule
\textbf{Llama3-70B}
& AWQ3 & -0.78 \\
& GPTQ3 & -0.81 \\

\bottomrule
\end{NiceTabular}
\end{center}
\caption{The final-layer residual magnitudes from the full-precision model have a strong negative correlation with the quantization errors of 3-bit methods.}
\label{table:uq_err_corr}
\end{table}

%% file: latex/Figures/density.tex
\begin{figure*}[!t]
\centering
\includegraphics[width=1\linewidth]{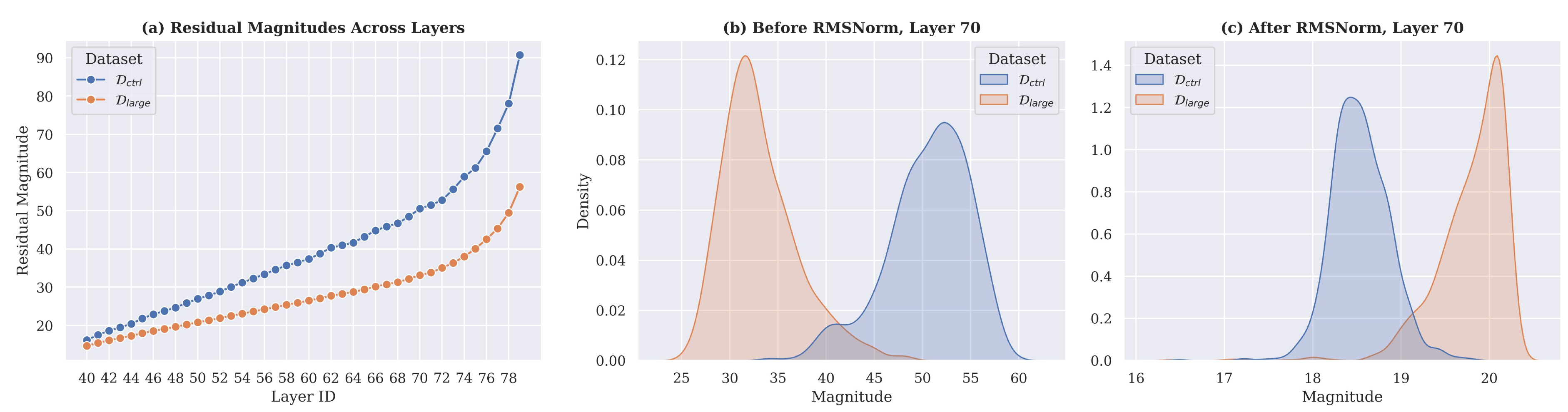}
\caption{(a) The residual magnitudes of the large-error set $\LargeD$ are smaller than those of $\Ctrl$ across layers. (b) \& (c) RMSNorm reverses the relative activation magnitudes. Before RMSNorm (b), the residual magnitudes of $\LargeD$ are smaller than $\Ctrl$, as shown in (a). After applying RMSNorm (c), the activation magnitudes of $\LargeD$ become larger. We observe clear reversal effects in the last 10 layers of the full-precision Llama3-70B.}
\label{fig:main-density}
\end{figure*}

%% file: latex/Tables/notations.tex
\begin{table}[!b]

\begin{center}
\begin{small}

\begin{tabular}{lll}
\toprule
\textbf{Notation} & & \textbf{Definition} \\
\midrule
$\Simple$ & $\RR^d$ & layer $l$ residual states (Eq.~\ref{eq:residual}) \\
$\PostLN$ & $\RR^d$ & RMSNorm$^{(l)}(\Simple)$ \\
$\Out$ & $\RR^d$ & MLP$^{(l)}(\PostLN)$ \\
$\gamma$ & $\RR^d$ & RMSNorm weights (Eq.~\ref{eq:rms})\\
$\Exnorm$ & $\RR$ & layer $l$ residual magnitude of $x$ \\
$\Norm (\Ds; \LLM; l)$ & $\RR^N$ & layer $l$ residual magnitudes of $\Ds$ \\
\bottomrule
\end{tabular}

\end{small}
\end{center}
\caption{Notation summary.}
\label{table:notation}

\end{table}

%% file: 06_where.tex
\section{Where Do Errors Arise From?}
\label{sec:where}
We study which parts of LLMs lead to large quantization errors with localization techniques: early exiting \cite{nostalgebraist,geva-etal-2022-transformer} and activation patching \cite{meng2022locating}.

\subsection{Early Exiting}
\label{sec:early_decode}
\citet{nostalgebraist} first introduces logit lens, an early exiting technique that projects hidden representations into the vocabulary space using the LLMs' output embeddings.
We use early exiting to measure the degree to which a residual state is ready for decoding.
Formally, let $\Simple$ be the residual state at layer $l$ and $\phi(\cdot)$ be the final RMSNorm operation applied before the output embedding matrix $U$.
We define the probability distribution from decoding $\Simple$ as:
\begin{equation*}
    \EE = \operatorname{Softmax} \left( U \cdot \phi(\Simple) \right)
\end{equation*}
Let $\EE_{\LLM}$ and $\EE_{\QLLM}$ be the early-exiting probabilities from the full-precision and quantized models.
We use $\EE_{\LLM}$ and $\EE_{\QLLM}$ to compute the average negative log-likelihood (NLL) over dataset $\Ds$ at each layer, denoted $\nll^{(l)}_{\LLM}(\Ds)$ and $\nll^{(l)}_{\QLLM}(\Ds)$, respecitvely.
A $\nll^{(l)}_\LLM(\Ds)$ value near the final model NLL means that the residual state at layer $l$ contains sufficient information for decoding; otherwise, further layers are needed to refine the output probability.

\input{latex/Tables/patch}
\input{latex/Figures/early-decode}
We investigate two key questions: (1) Can we observe patterns from the early exiting dynamics that distinguish $\LargeD$ and $\Ctrl$? (2) At which layer do $\nll^{(l)}_{\LLM}(\Ds)$ and $\nll^{(l)}_{\QLLM}(\Ds)$ begin to diverge?
To answer these questions, we perform four runs of early exiting, one for each combination of model type 
(full-precision and quantized) and dataset split ($\LargeD$ and $\Ctrl$). 
We then compute $\nll^{(l)}_{\LLM}(\LargeD)$, $\nll^{(l)}_{\QLLM}(\LargeD)$, $\nll^{(l)}_{\LLM}(\Ctrl)$, and $\nll^{(l)}_{\QLLM}(\Ctrl)$ across the upper half layers.

Fig.~\ref{fig:main-early} compares how the four NLL values evolve across the upper layers of Mistral-12B, where the quantized model $\QLLM$ is produced by AWQ3.
We observe a clear distinction between $\LargeD$ and $\Ctrl$ on the full-precision model: $\LargeD$ (purple) has higher NLLs than $\Ctrl$ (green) until layer 36, around which its NLLs begin to decrease sharply.
This observation supports our claim in \cref{sec:destiny} that the properties of the full-precision model reflect the future quantization errors. 
Moreover, the sharp NLL decrease in later layers suggests that $\LargeD$ heavily relies on these layers to make accurate predictions.
This reliance requires the quantized model to keep the residual states precise through to the later layers, which is a challenging task because quantization errors propagate over layers.
Otherwise, the model cannot decode the noisy representations, resulting in large quantization errors.
By comparison, $\Ctrl$ relies less on the later layers, and the quantized model (cyan) shows less deviation from the full-precision one (green).

An alternative hypothesis is that the precision of weights in the later layers is crucial to $\LargeD$.
To test this hypothesis, we keep all the weights above layer 32 in full-precision, because Fig.~\ref{fig:main-early} shows that on $\LargeD$, the quantized model (orange) starts to diverge from the full-precision model (purple) at layer 33.
This experiment restores 7 layers ($18\%$) from 3 bits to 16 bits.
However, the perplexity on $\LargeD$ only slightly decreases from $8.67$ to $8.05$, versus $5.60$ for the full-precision model.
We show similar results on other LLMs in \cref{sec:app_ee}.

In summary, our early exiting experiment suggests that various methods struggle with $\LargeD$ because it requires the quantized models to maintain precise residual states up to the late layers.
We also find that leaving late layers unquantized has little effect on NLLs, implying that the cumulative error in residual states, not the precision of upper-layer weights, is the primary cause for large errors.

\subsection{Cross-Model Activation Patching}
\label{sec:patch}
\citet{prakash2023fine} introduce cross-model activation-patching (CMAP) to identify how fine-tuned models improve over the pretrained ones.
We adapt CMAP to track where the quantized models differ from the full-precision ones, aiming to identify the sources of quantization errors. 
For a given example from $\LargeD$, we first run a forward pass on the full-precision model and cache its activations. 
We then run the same example through the quantized model, but replace the activations of specific modules with those from the full-precision model.
We patch the outputs of MHA, denoted $\HAttn$, and the outputs of different modules in MLP.
Specifically, all LLMs we study implement MLP with Gated Linear Units \cite{glu}:
\begin{align*}
     & \HGate = \sigma(\Gate \; h), \\
     & \HUp = \Up \; h, \\
     & \HDown = \Down \; (\HGate \odot \HUp),
     \label{eq:glu}
\end{align*}
where $h$ is the post-RMSNorm input to MLP, $\Gate$, $\Up$, and $\Down$ are the MLP weights, $\HDown$ is the output of MLP, and $\sigma(\cdot)$ is the activation function.
To trace quantization errors, we perform four runs of CMAP, corresponding to patching $\HGate$, $\HUp$, $\HDown$, and $\HAttn$, in the upper half layers of the models.
Patching $\HDown$ is a much stronger intervention than patching either $\HUp$ or $\HDown$, as it overrides the final outputs of MLPs.

We apply AWQ3 and NF3 methods and report the perplexity on $\LargeD$ in Table~\ref{table:patch}.
First, we find that patching the outputs of MLP $\HDown$ greatly narrows the perplexity gaps between the 3-bit and 16-bit full-precision models, bringing the gaps below $0.5$ across all LLMs.
In contrast, patching the outputs of MHA $\HAttn$ only leads to marginal effects, suggesting that the quantization errors from MLPs are dominating the results.
We further find that inside MLPs, patching $\HUp$ alone is ineffective, but patching $\HGate$ results in substantial perplexity reduction, approaching the effect of patching the entire MLP outputs $\HDown$. 
These results highlight the importance of having precise outputs from MLP gates, which control the information passed on to subsequent layers.

We conduct two additional experiments to understand the source of quantization errors in MLP gates: (1) only patching the inputs to $\Gate$, and (2) restoring all $\Gate$ weights to 16-bit precision. 
We find that patching the gates' inputs (1) performs nearly as well as patching the outputs, $\HGate$, while restoring the weights (2) is ineffective. 
For instance, on Llama-3-8B quantized with NF3, (1) achieves a perplexity of 6.53 compared to 10.45 for (2). 
These results further support our conclusion in \cref{sec:early_decode} that noisy activations are the primary source of large quantization errors.

%% file: latex/Tables/patch.tex
\begin{table*}[!t]
\begin{center}
\begin{small}
\begin{NiceTabular}{lccccccc}
    \CodeBefore
    \rectanglecolor{bggreen}{1-1}{8-2}
    \rectanglecolor{bgblue}{1-3}{8-4}
    \rectanglecolor{bgyellow}{1-5}{8-9}
    \Body
\toprule
\textbf{Model} & \textbf{Method} & \textbf{16-bit} & \textbf{3-bit} & \textbf{Patch} \boldmath{$\HDown$} & \textbf{Patch} \boldmath{$\HGate$} & \textbf{Patch} \boldmath{$\HUp$} & \textbf{Patch} \boldmath{$\HAttn$} \\
\midrule
Qwen2.5-7B 
& AWQ3 & \multirow{2}{*}{$9.42$} & $12.26$ \scriptsize{(+$2.84$)} & $9.73$ \scriptsize{(+$0.31$)} & $10.20$ \scriptsize{(+$0.78$)} & $12.64$ \scriptsize{(+$3.22$)} & $11.12$ \scriptsize{(+$1.70$)} \\
& NF3 &  & $12.53$ \scriptsize{(+$3.11$)} & $9.66$ \scriptsize{(+$0.24$)} & $10.61$ \scriptsize{(+$1.19$)} & $12.97$ \scriptsize{(+$3.55$)} & $11.35$ \scriptsize{(+$1.93$)} \\

\midrule 
Llama3-8B 
& AWQ3 & \multirow{2}{*}{$5.62$} & $10.68$ \scriptsize{(+$5.06$)} & $5.95$ \scriptsize{(+$0.33$)} & $6.42$ \scriptsize{(+$0.80$)} & \phantom{1}$9.97$ \scriptsize{(+$4.35$)} & $9.27$ \scriptsize{(+$3.65$)} \\
& NF3 &  & $11.02$ \scriptsize{(+$5.40$)} & $5.92$ \scriptsize{(+$0.30$)} & $6.43$ \scriptsize{(+$0.81$)} & $10.48$ \scriptsize{(+$4.86$)} & $9.49$ \scriptsize{(+$3.87$)} \\
\midrule
Mistral-12B 
& AWQ3 & \multirow{2}{*}{$5.60$} & \phantom{1} $8.67$ \scriptsize{(+$3.07$)} & $6.05$ \scriptsize{(+$0.45$)} & $6.13$ \scriptsize{(+$0.53$)} & $12.49$ \scriptsize{(+$6.89$)} & $7.61$ \scriptsize{(+$2.01$)} \\
& NF3 &  & $10.14$ \scriptsize{(+$4.54$)} & $5.92$ \scriptsize{(+$0.32$)} & $7.12$ \scriptsize{(+$1.52$)} & $13.75$ \scriptsize{(+$8.15$)} & $8.71$ \scriptsize{(+$3.11$)} \\
\midrule
Llama3-70B & AWQ3 & $2.66$ & \phantom{1}$5.25$ \scriptsize{(+$2.59$)} & $2.73$ \scriptsize{(+$0.07$)} & $2.94$ \scriptsize{(+$0.28$)} & $3.24$ \scriptsize{(+$0.58$)} & $4.73$ \scriptsize{(+$2.07$)} \\

\bottomrule
\end{NiceTabular}
\end{small}
\caption{The perplexity ($\downarrow$) on $\LargeD$ before/after patching activations from 16-bit full-precision models into 3-bit models, quantized with AWQ3 and NF3 methods, respectively. Both patching the entire MLP outputs ($\HDown$) and patching only the MLP gate outputs ($\HGate$) can substantially close the gap between the 3-bit and 16-bit models.}
\label{table:patch}
\end{center}
\end{table*}

%% file: latex/Figures/early-decode.tex
\begin{figure}[!t]
\centering
\includegraphics[width=1.\linewidth]{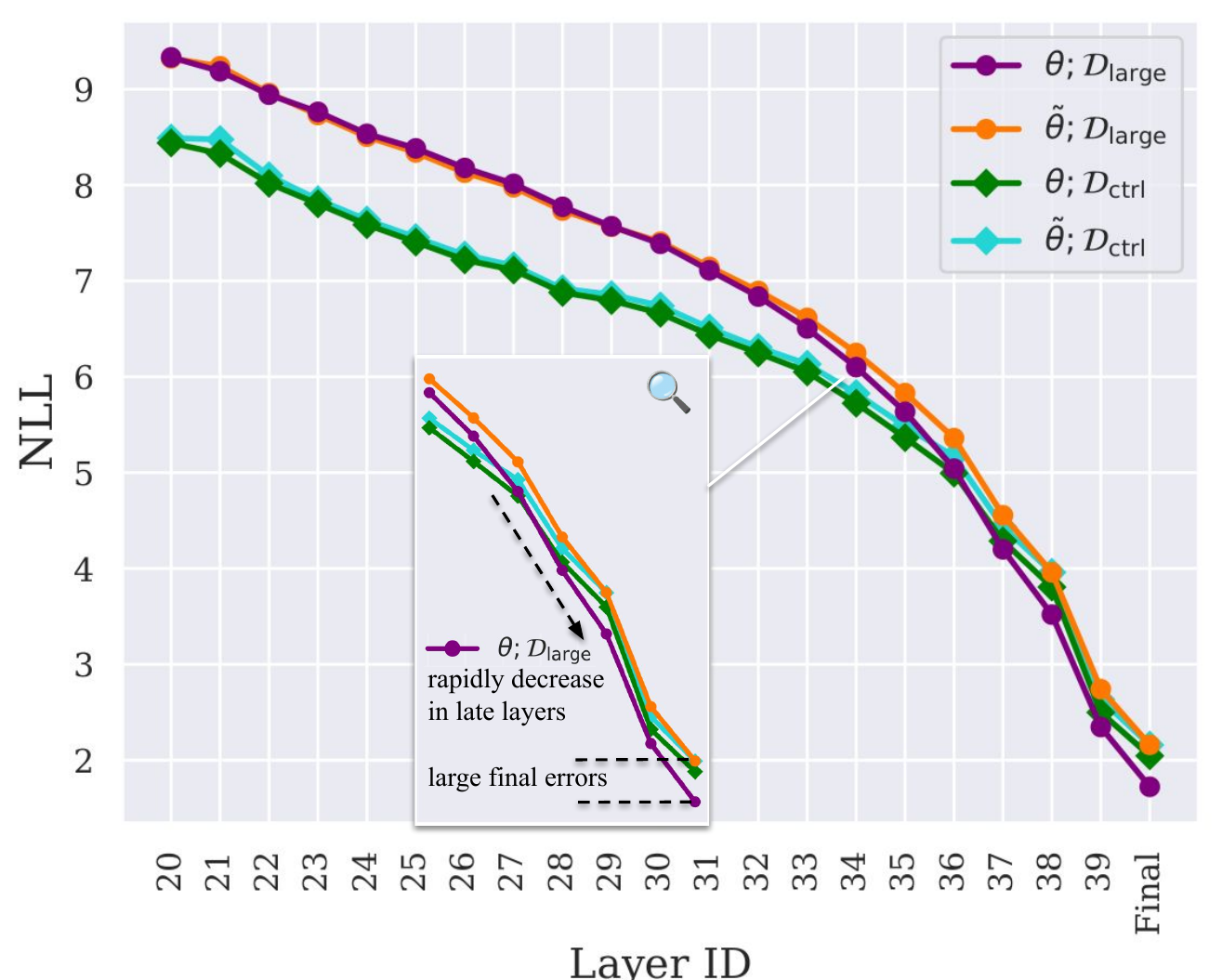}
\caption{Early exiting across layers shows that the NLLs of $\LargeD$ (purple and orange) dramatically decrease in the late layers, suggesting that $\LargeD$ heavily relies on the late layers to adjust the output probabilities. Meanwhile, the quantized model (orange) does not show obvious deviation from the full-precision one (purple) until layer 33.}
\label{fig:main-early}
\end{figure}

%% file: 07_data_analysis.tex
\section{What Data Cause Large Errors?}
Understanding what kinds of data lead to large quantization errors is crucial for future improvement.
This section studies whether examples in $\LargeD$ cluster in certain domains or are unusual, long-tail data. 

\subsection{Categorizing Examples in $\LargeD$}
To better understand the distribution of $\LargeD$, we apply the topic classifier from \citet{wettig2025organize}, which categorizes web pre-training data into 24 topics.\footnote{ \url{https://huggingface.co/WebOrganizer/TopicClassifier-NoURL}}
We find that the $\LargeD$ sets of both Mistral-12B and Llama3-70B cover all the 24 topics, and we plot the topic distributions in Fig.~\ref{fig:topic}.
For both models, \texttt{Entertainment}, \texttt{Finance \& Business}, \texttt{Politics}, and \texttt{Health} are among the top five topics, although none account for more than $12\%$ of the dataset.
We also manually inspect $\LargeD$ examples and do not find them unusual.

\subsection{Quantization Errors vs. Long-Tail Data}
\label{sec:data_analysis}
\citet{ogueji-etal-2022-intriguing,marchisio2024-quantization} show that model compression has a disparate effect for long-tail data; thus, we explore the relationship between quantization errors and data long-tailness.

\paragraph{FineWeb.} 
We use the data likelihood assigned by a strong language model, Llama3-70B, to estimate the long-tail characteristics of the examples.
We call the FineWeb examples that have small log-likelihoods (large NLLs) on the full-precision Llama3-70B as long-tail examples.
Fig.~\ref{fig:rare} shows that long-tail examples (x-axis $\ge 4$) have small quantization errors under both AWQ3 (avg. $0.14$) and GPTQ3 (avg. $0.30$) methods.
On the other hand, examples that suffer from large errors have widespread NLLs before quantization, 
and many ``head data" (small NLLs; x-axis $\le 1$) have large quantization errors (avg. $0.67$ for AWQ3 and $1.00$ for GPTQ3).
We observe the same findings on the other LLMs (see Fig.~\ref{fig:app-rare} in appendix).
\input{latex/Figures/topic}
\paragraph{PopQA.} 
We investigate how quantization affects LLMs' factual knowledge on entities with different popularity levels.
We partition the PopQA dataset into buckets based on the examples' $\log_{10} (\text{popularity})$ and calculate the accuracies within each bucket.
Fig.~\ref{fig:main-popqa} presents the accuracy degradation after quantization under different buckets, where we apply NF3 (blue) and AWQ3 (orange) to quantize Mistral-12B into 3 bits.
We find that when the examples have the lowest and highest popularities (the leftmost and rightmost buckets, respectively), the 3-bit models show the lowest degradation.
In contrast, examples in the middle buckets (log popularity between $2.8$ and $4.8$) are most affected by quantization, with accuracies dropped by $> 10\%$.
Note that only $6\%$ of examples have log popularity greater than $4.8$, meaning that quantization can severely degrade LLMs' knowledge on entities that have intermediate to high popularity.
We include results from other LLMs in Fig.~\ref{fig:app-popqa-delta}, showing that low popularity examples usually have milder degradation.

\paragraph{Summary.}
We find that long-tail examples are relatively less affected by quantization on both FineWeb and PopQA. 
Meanwhile, well-represented data are not immune to large errors.

%% file: latex/Figures/topic.tex
\begin{figure}[!t]
\centering
\includegraphics[width=1.0\columnwidth]{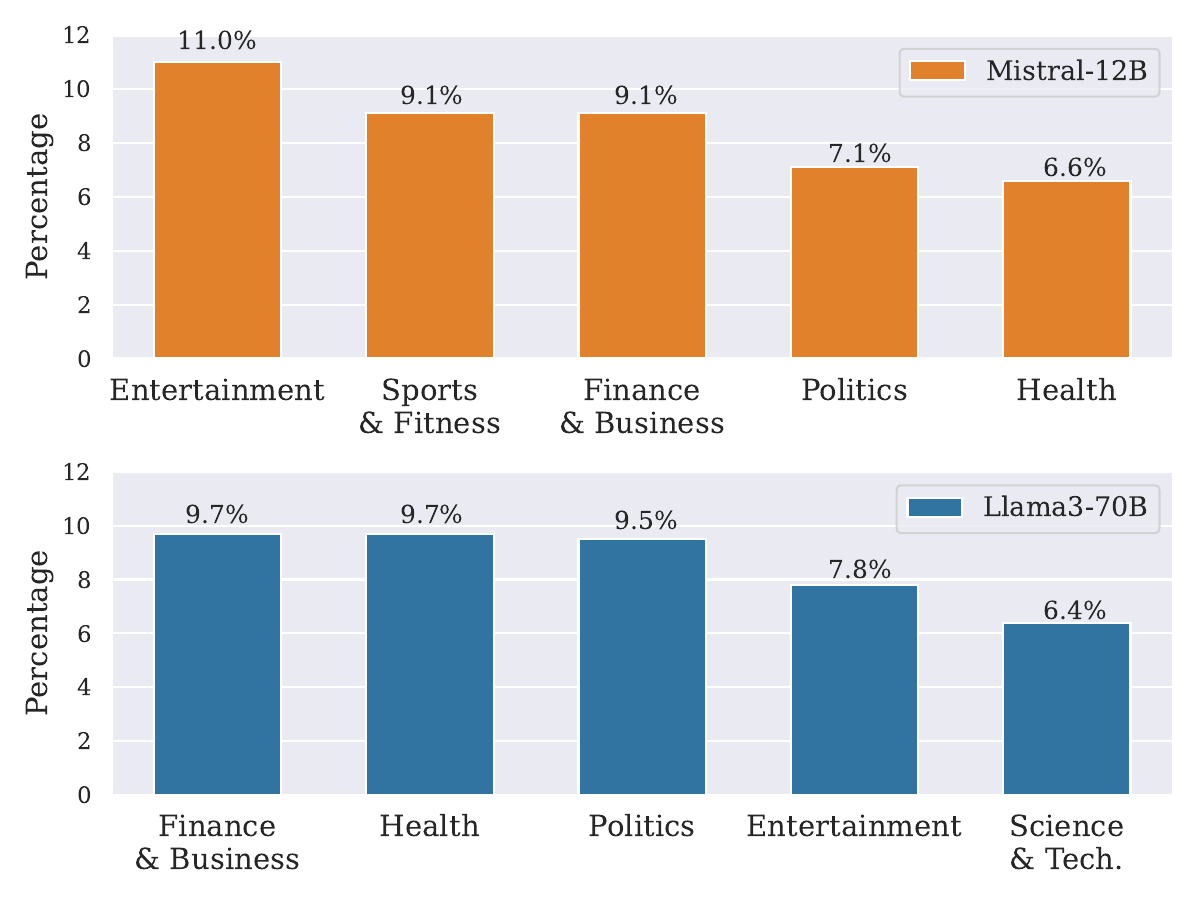}
\caption{Top five topics of $\LargeD$ for Mistral-12B (up; orange) and Llama3-70B (down; blue), respectively.}
\label{fig:topic}
\end{figure}

%% file: 08_conclusion.tex
\section{Discussion}
We thoroughly study why certain examples are disproportionately affected by 3--4 bit, weight-only quantization methods.
We reveal the distinct nature of large-error examples before quantization, showing that these examples have smaller residual magnitudes and rely on upper layers to refine output probabilities.
Our patching and weight recovery experiments further highlight the importance of having precise activations at MLP gates, and reject naive mixed-precision approaches.
Our kurtosis value analysis finds that large-error examples actually have fewer outliers in residual states across upper layers.
We further discover the reversal effects of RMSNorm, showing that large quantization errors may stem from overall larger activations across dimensions, rather than from a few outlier dimensions.
Finally, we explore the relationship between quantization errors and long-tail data, showing that long-tail examples tend to have less degradation after quantization than other examples.

\input{latex/Figures/rare}
Overall, our work sheds light on why and where large quantization errors occur.
Our findings may inspire quantization error prediction frameworks and encourage future post-training methods or quantization-friendly architectures that specifically address MLP gates and RMSNorm; for instance, by avoiding the Gated Linear Units or redesigning the LayerNorm \cite{Zhu2025DyT}.

\input{latex/Figures/popqa-bar}

%% file: latex/Figures/rare.tex
\begin{figure}[!t]
\centering
\includegraphics[width=0.9\columnwidth]{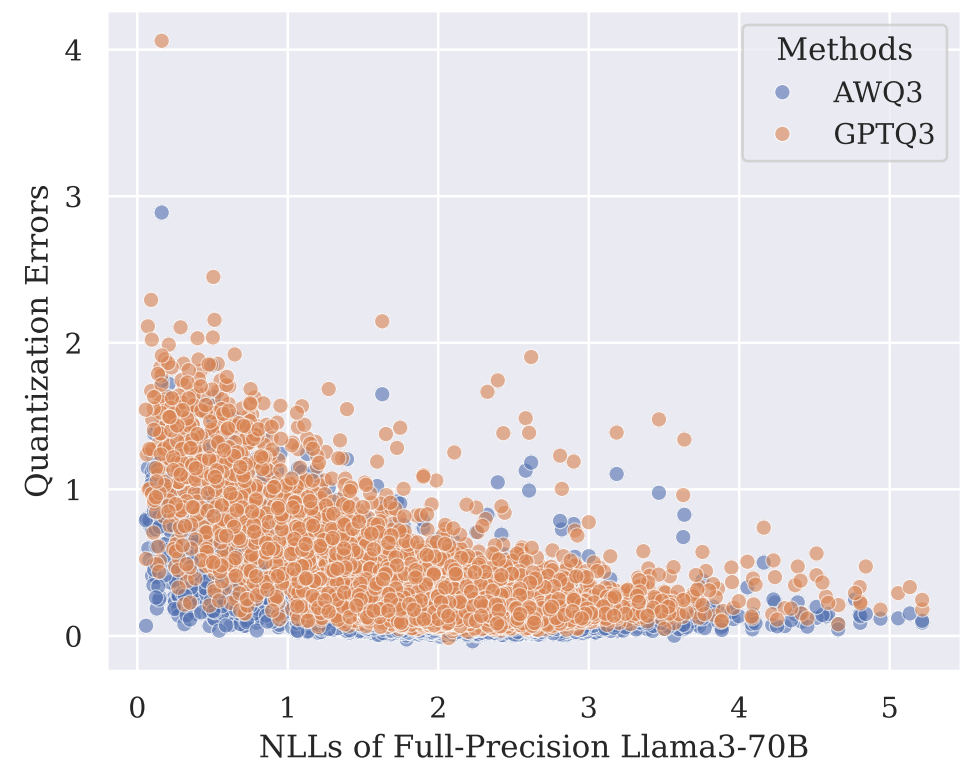}
\caption{Long-tail samples with large NLLs on the full-precision Llama3-70B have small quantization errors. Each dot represents a FineWeb example. We quantize Llama3-70B with AWQ3 and GPTQ3 methods and compute their quantization errors with Eq.~\ref{eq: error}.}
\label{fig:rare}
\end{figure}

%% file: latex/Figures/popqa-bar.tex
\begin{figure}[!t]
\centering
\includegraphics[width=1.\columnwidth]{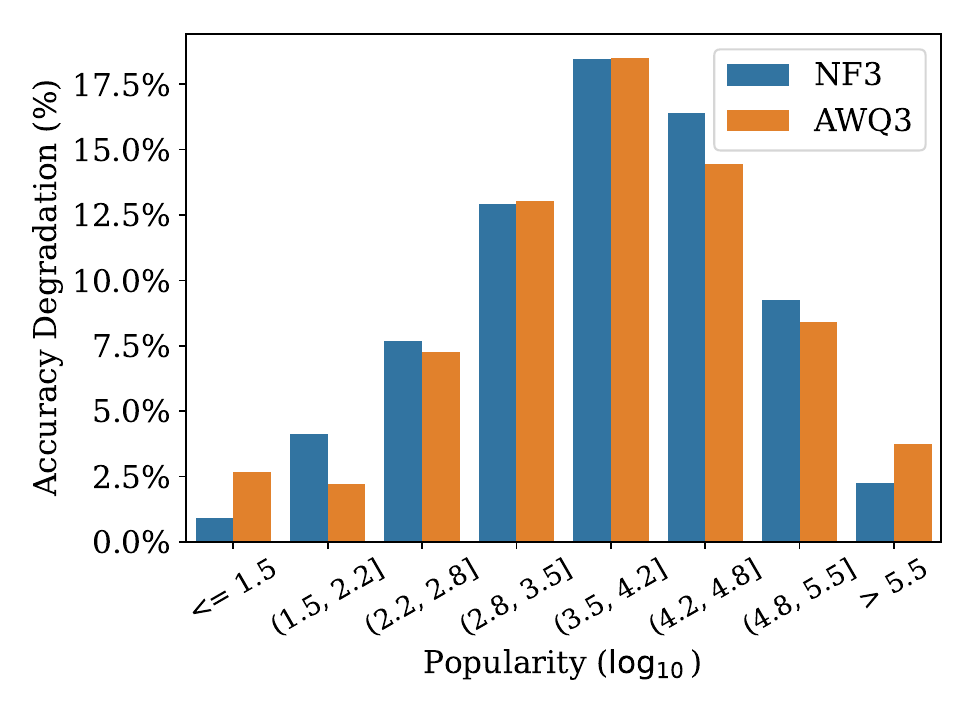}
\caption{The accuracy degradation of PopQA under different levels of popularity after 3-bit quantization. Samples with log popularity between $2.8$ and $4.8$ show the greatest degradation. }
\label{fig:main-popqa}
\end{figure}

%% file: 09_limitations.tex
\section{Limitations}
We conduct our main experiments on FineWeb and use NLL as our evaluation metric, which does not consider downstream tasks.
We believe this is a reasonable choice because \citet{jin2024comprehensive} show that perplexity (exponentiated average NLL) is a reliable performance indicator for quantized LLMs on various evaluation benchmarks.
We invite future work beyond intrinsic analysis to extrinsic, downstream tasks.

We do not cover sub-3-bit settings because we observe that the methods in our study suffer from extremely large errors when quantized below 3 bits (with perplexity greater than $10^3$ and near-random accuracy on commonsense QA tasks), making it hard to pinpoint the cause of errors.
Because prior work also marks 3 bits as a turning point in quantization \cite{dettmers2023case,liu2025paretoq}, our study of quantization errors under the 3--4 bit regime could be valuable to future efforts to improve quantization methods to lower bits.

%% file: acknowledgements.tex
\section*{Acknowledgements}
We thank Yuqing Yang for helpful discussions, Ming Zhong for guidance on table coloring, and the anonymous reviewers for their valuable feedback.
This work was supported in part by the National Science Foundation under Grant No. IIS-2403436. Any opinions, findings, and conclusions or recommendations expressed in this material are those of the author(s) and do not necessarily reflect the views of the National Science Foundation.

%% file: 10_appendix.tex
\label{sec:appendix}

\section{Implementation Details}
\label{sec:app_details}
We aim to answer whether different quantization methods make similar errors; thus, we follow the original implementations and do not unify their quantization configurations.
\paragraph{GPTQ.} We use the \texttt{GPTQModel}\footnote{\url{https://github.com/ModelCloud/GPTQModel}} package for implementation.
The calibration set consists of samples from English C4.\footnote{\url{en/c4-train.00001-of-01024.json.gz}} The quantization group size is 64 for 3-bit models and 128 for 4-bit models.
We do not include the results of GPTQ3 with Llama-3-8B because we observe severe perplexity degradation with the latest \texttt{GPTQModel} package.

\paragraph{AWQ.} We follow the original implementation\footnote{\url{https://github.com/mit-han-lab/llm-awq}} and use the released model checkpoints when available.
The calibration set consists of samples from the Pile \cite{gao2020pile}.
The quantization group size is 128 for both 3- and 4-bit models.
Note that because AWQ applies scaling transformation, in our patching experiments \cref{sec:patch}, we also apply the same scaling factors on the \emph{full-precision} models, which does not change their outputs but aligns their activations with the AWQ-quantized models.

\paragraph{NF.} We use the official packages: \texttt{bitsandbytes}\footnote{\url{https://pypi.org/project/bitsandbytes}} for 4-bit models and \texttt{flute}\footnote{\url{https://github.com/HanGuo97/flute}} for 3-bit model.
The quantization group size is 64 for both 3- and 4-bit models.
We do not have the results of NF3 on Llama-3-70B due to kernel incompatibility.

\paragraph{EQAT.} EQAT train on thousands of samples from RedPajama \cite{together2023redpajama}. We use the officially released 3-bit model checkpoints,\footnote{\url{https://github.com/OpenGVLab/EfficientQAT}} which only cover the Llama3 models.
The quantization group size is 128.
\paragraph{Perplexity.} We report the perplexity of different 3- and 4-bit quantization methods in Table~\ref{table:app_ppl_llama3-8}.
We follow the standard implementation of WikiText perplexity that sets the sequence length to 2048. The sequence length of FineWeb is 512.

\input{latex/Tables/app_ppl_llama3-8}
\paragraph{Data.}
We use FineWeb for our main experiments because it is a diverse, cleaned dataset, and none of the quantization methods considered above sample calibration data from it, making FineWeb an unbiased corpus for our study.
We also explore more technical domains, Arxiv and OpenWebMath~\cite{azerbayev2023llemma,paster2023openwebmath}.
Our findings that the quantization errors of different methods are strongly correlated (\cref{sec:diff_q}) also hold on these domains.
Specifically, the average correlation across 8 pairs of methods on Llama3-70B is 0.69 on ArXiv and 0.89 on OpenWebMath.
\paragraph{Definition of Error.}
We also explore an alternative definition of quantization errors based on KL divergence. 
Let $p_\LLM$, the probability distribution of the full-precision model, be the true distribution.
We can define errors as the KL divergence between the quantized and full-precision model, $\mathrm{KL}\left( p_\theta \,\|\, p_{\QLLM} \right)$.
The results of these two definitions are highly correlated ($\corr \ge 0.97$).
Thus, we adopt Eq.~\ref{eq: error} for its lower computation costs.

\section{The Construction of the $\LargeD$}
\label{sec:app_dlarge}
To ensure the diversity of $\LargeD$ examples, we apply aggressive data deduplication based on Jaccard similarity \cite{jaccard1912distribution}, filtering out examples with 5-gram Jaccard similarity $> 0.1$.
We create a separate $\LargeD$ for each quantization method and observe similar findings across them.
Besides, the $\LargeD$ sets of different methods under the same LLMs often have overlapping examples, with the average Jaccard similarity $\frac{|A \cap B|}{|A \cup B|} = 0.30, 0.49, 0.66, 0.66$ on Qwen2.5-7B, Meta-Llama-3-8B, Mistral-12B, and Meta-Llama-3-70B, respectively.
Therefore, in the main content, we always use the one derived from AWQ3 as $\LargeD$ for simplicity.

\section{Correlations Between Residual Magnitudes and Quantization Errors}
\label{sec:app_corr_res_err}
We present the correlations between the quantization errors $\EDs(\Ds ; \QLLM)$ and the residual magnitudes $\Norm (\Ds; \LLM; l)$ across layers of Qwen2.5-7B, Llama3-8B, Mistral-12B, and Llama3-70B in Fig.~\ref{fig:app-norm-Qwen2.5-7b}, \ref{fig:app-norm-llama3-8b}, \ref{fig:app-norm-mistral-12b}, \ref{fig:app-norm-llama3-70b}, respectively.
We find that different models and methods all show that the correlations become increasingly negative in deeper layers.

\section{RMSNorm and Outliers}
\label{sec:app_rms_outlier}
First, we select the serverest outlier dimensions in $\Simple$, namely, the most-activated dimensions in $\Simple$ that have the highest average absolute activation over the 10k FineWeb dataset.
Next, we rank the entries in RMSNorm weights $\gamma$ by their absolute values.
We compare the median of $\gamma$ with those entries corresponding to the outlier dimensions. 
Table~\ref{table:app_gamma_stats} shows that in the upper half layers of Llama3-70B, all outlier dimensions have much smaller weights than the median.
On the other hand, in Mistral-12B, the most severe outliers have small weights (rank 1 or 2) across layers, but other outlier dimensions sometimes have much larger weights than the median.
These results show that RMSNorm weights suppress the largest outlier dimensions but sometimes also promote other outlier dimensions.
In Fig.~\ref{fig:app-kappa} and Fig.~\ref{fig:app-kappa-af}, we show the kurtosis values before and after RMSNorm, respectively.
Comparing the scale of the kurtosis values (y-axis) of the two figures, we find that the scale after RMSNorm is much smaller, implying that RMSNorm in the upper layers suppresses outliers overall.

Similar to \citet{wei2022outlier}, we find that the LayerNorm parameters modulate activation outliers; however, we observe mostly opposite effects, likely due to the different architectural designs (pre-LN vs. post-LN \citealp{xiong2020layer}).

\input{latex/Tables/corr_app}
\input{latex/Tables/app_gptq_variants}
\section{More Early Exiting Results}
\label{sec:app_ee}
We apply early exiting on $\MLP$ and $\Simple$, the hidden states after the residual operations (see Eq.~\ref{eq:residual_mlp}), across layers.
Figure \ref{fig:app-ee-qwen}, \ref{fig:app-ee-llama3-8b}, \ref{fig:app-ee-mistral}, and ~\ref{fig:app-ee-llama3-70b} shows the results on Qwen2.5-7B, Llama3-8B, Mistral-12B, and Llama3-70B, respectively, where we use AWQ3 to produce the quantized models.
All the LLMs show that the NLLs of $\LargeD$ (purple) decrease dramatically in the later layers, suggesting that $\LargeD$ relies more on the late layers to decode the representations than $\Ctrl$ (green).

\input{latex/Figures/app-kappa}
\input{latex/Figures/app-kappa-af-ln}
\input{latex/Figures/app-corr-norms-Qwen2.5-7B}
\input{latex/Figures/app-corr-norms-Meta-Llama-3-8B}
\input{latex/Figures/app-corr-norms-Mistral-Nemo-Base-2407}
\input{latex/Figures/app-corr-norms-Meta-Llama-3-70B}

\input{latex/Figures/app-ee-Qwen2.5-7B}
\input{latex/Figures/app-ee-Meta-Llama-3-8B}
\input{latex/Figures/app-ee-Mistral-Nemo-Base-2407}
\input{latex/Figures/app-ee-Meta-Llama-3-70B}
\input{latex/Figures/app-rare}
\input{latex/Figures/app-popqa-delta}
\input{latex/Figures/app-popqa-abs}
\input{latex/Figures/app-popqa-data}

\section{Full Results on PopQA}
\label{sec:app_popqa}
Fig.~\ref{fig:app-popqa-delta} and \ref{fig:app-popqa-abs} show the accuracy degradation and absolute accuracy values of different models, respectively.
Fig.~\ref{fig:app-popqa-data} shows the popularity histogram.

\input{latex/Tables/app_gamma_stats}

%% file: latex/Tables/app_ppl_llama3-8.tex
\begin{table}[!t]
\begin{center}
\begin{small}
\begin{tabular}{llcc}
\toprule
Model & Method  &    Fineweb   & Wiki \\
\midrule
Qwen2.5-7B 
& 16 bits    &   11.06  &  6.85 \\
& AWQ4       &   11.35  &  7.09 \\ 
& NF4        &   11.37  &  7.10 \\
& GPTQ4      &   11.33  &  7.16 \\
& AWQ3       &   12.77  &  8.20 \\
& NF3        &   13.47  &  8.51 \\
& GPTQ3      &   12.26  &  8.11 \\
\midrule
Llama3-8B 
& 16 bits    &   9.57  &  6.14 \\
& AWQ4       &   10.13  &  6.56 \\
& NF4        &   10.11  &  6.56 \\
& GPTQ4      &   10.12  &  6.86 \\
& AWQ3       &   12.13  &  8.19 \\
& NF3        &   12.45  &  8.44 \\
& EQAT3      &   10.77  &  7.10 \\
\midrule
Mistral-12B 
& 16 bits      &      8.67   & 5.75 \\
& AWQ4         & 9.09  &  6.01 \\
& NF4          &  9.13 &   6.06 \\
& GPTQ4        &  9.02 &   6.04 \\
& AWQ3         & 10.32 &   7.03 \\
& NF3          & 12.48 &   8.75 \\
& GPTQ3        & 10.44 &   7.20 \\
\midrule
Llama3-70B 
& 16 bits       & 7.15 &   2.86 \\
& AWQ4          & 7.42  &  3.27 \\
& NF4           & 7.88  &  3.22 \\
& GPTQ4         & 7.52  &  3.58 \\
& GPTQ3         & 9.64  &  6.51 \\
& AWQ3          & 8.40   &  4.74 \\

\bottomrule
\end{tabular}
\end{small}
\end{center}
\caption{The perplexity of different quantization methods on the 10k FineWeb dataset and WikiText. Our FineWeb examples have shorter sequence lengths than WikiText (512 vs. 2048), leading to higher perplexity.}
\label{table:app_ppl_llama3-8}
\end{table}

%% file: latex/Tables/corr_app.tex
\begin{table}[!t]
\begin{center}
\begin{small}
\begin{tabular}{llc}
\toprule
Model & (Method1, Method2) & $\corr (\EDs_1, \EDs_2)$ \\
\midrule
Qwen2.5-7B 
& ( AWQ4 , NF4   ) & 0.52 \\
& ( AWQ4 , GPTQ4 ) & 0.66 \\
& ( AWQ4 , NF3   ) & 0.61 \\
& ( AWQ4 , GPTQ3 ) & 0.65 \\
& ( AWQ3 , NF3   ) & 0.78 \\
& ( AWQ3 , GPTQ3 ) & 0.84 \\
& ( NF4  , GPTQ4 ) & 0.57 \\
& ( NF4  , AWQ3  ) & 0.57 \\
& ( NF4  , GPTQ3 ) & 0.55 \\
& ( NF3  , GPTQ3 ) & 0.75 \\
& ( GPTQ4, AWQ3  ) & 0.75 \\
& ( GPTQ4, NF3   ) & 0.70 \\

\midrule
Llama3-8B 
& ( AWQ4 , NF4   ) & 0.95 \\
& ( AWQ4 , GPTQ4 ) & 0.96 \\
& ( AWQ4 , NF3   ) & 0.88 \\
& ( AWQ4 , EQAT3 ) & 0.92 \\
& ( AWQ3 , NF3   ) & 0.98 \\
& ( AWQ3 , EQAT3 ) & 0.97 \\
& ( NF4  , GPTQ4 ) & 0.95 \\
& ( NF4  , AWQ3  ) & 0.87 \\
& ( NF4  , EQAT3 ) & 0.90 \\
& ( NF3  , EQAT3 ) & 0.96 \\
& ( GPTQ4, AWQ3  ) & 0.90 \\
& ( GPTQ4, NF3   ) & 0.90 \\
& ( GPTQ4, EQAT3 ) & 0.93 \\
\bottomrule
\end{tabular}
\end{small}
\end{center}
\caption{The quantization errors of different methods are strongly correlated on the FineWeb. The numbers in the method names denote 3 or 4-bit precision.}
\label{table:app_fineweb_corr}
\end{table}

%% file: latex/Tables/app_gptq_variants.tex
\begin{table}[!t]
\begin{center}
\begin{tabular}{lc}
\toprule
\textbf{Variant Pairs} & \boldmath{$\rho$} \\
\midrule
(damp-0.005-sort, damp-0.01-sort) & 0.95 \\
(damp-0.005-sort, damp-0.02-sort) & 0.95 \\
(damp-0.01\phantom{1}-sort, damp-0.02-sort)  & 0.95 \\
(damp-0.005-sort, damp-0.01-no\_sort) & 0.92 \\
(damp-0.01\phantom{1}-sort, damp-0.01-no\_sort) & 0.92 \\
(damp-0.02\phantom{1}-sort, damp-0.01-no\_sort) & 0.92 \\
\bottomrule
\end{tabular}
\end{center}
\caption{Quantization error correlations of 3-bit Qwen2.5-7B GPTQ variants with different damping coefficients and activation-sorting settings. All pairs are strongly correlated.}
\label{table:app_gptq}
\end{table}

%% file: latex/Figures/app-kappa.tex
\begin{figure*}[t!]
    \centering
    \includegraphics[width=1.\linewidth]{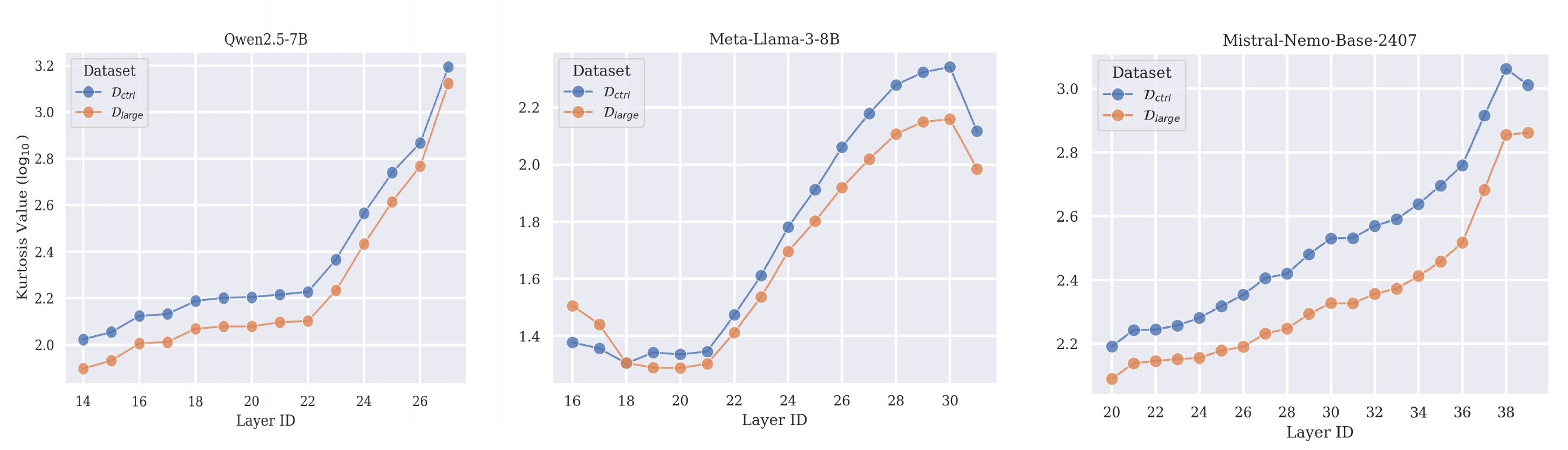}
    \caption{Pearson's kurtosis values of the residual states $\Simple$ from different full-precision LLMs (log scale). The large-error set $\LargeD$ shows lower kurtosis values than the control set $\Ctrl$ across layers, indicating that $\LargeD$ contains fewer or less extreme outliers in the residual states.}
    \label{fig:app-kappa}
\end{figure*}

%% file: latex/Figures/app-kappa-af-ln.tex
\begin{figure*}[t!]
    \centering
\includegraphics[width=1.\linewidth]{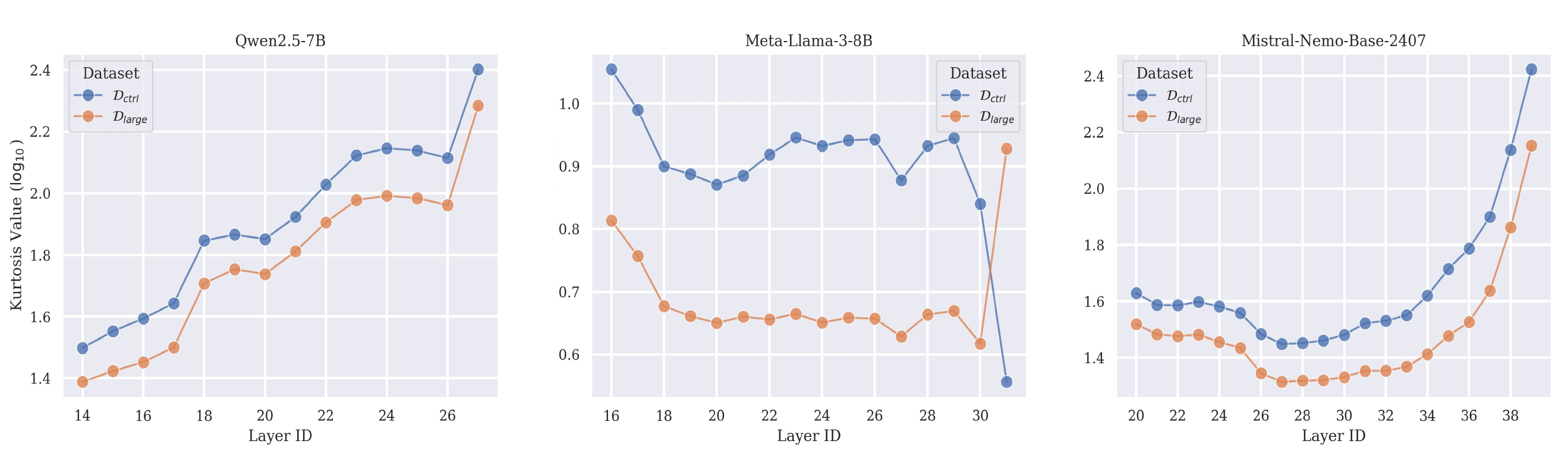}
\caption{Pearson's kurtosis values of the post-RMSNorm hidden states $\PostLN$ from different full-precision LLMs (log scale). The large-error set $\LargeD$ shows lower kurtosis values than the control set $\Ctrl$, except for the last layer of Llama3-8B, indicating that overall $\LargeD$ contains fewer or less extreme outliers in the hidden states after RMSNorm. In addition, compared with Fig.~\ref{fig:app-kappa} (before RMSNorm), the scale of the kurtosis values in this figure is much smaller (y-axis), implying that RMSNorm in the upper layers tends to suppress outliers.}
    \label{fig:app-kappa-af}
\end{figure*}

%% file: latex/Figures/app-corr-norms-Qwen2.5-7B.tex
\begin{figure*}[t!]
    \centering
    \includegraphics[width=1.\linewidth]{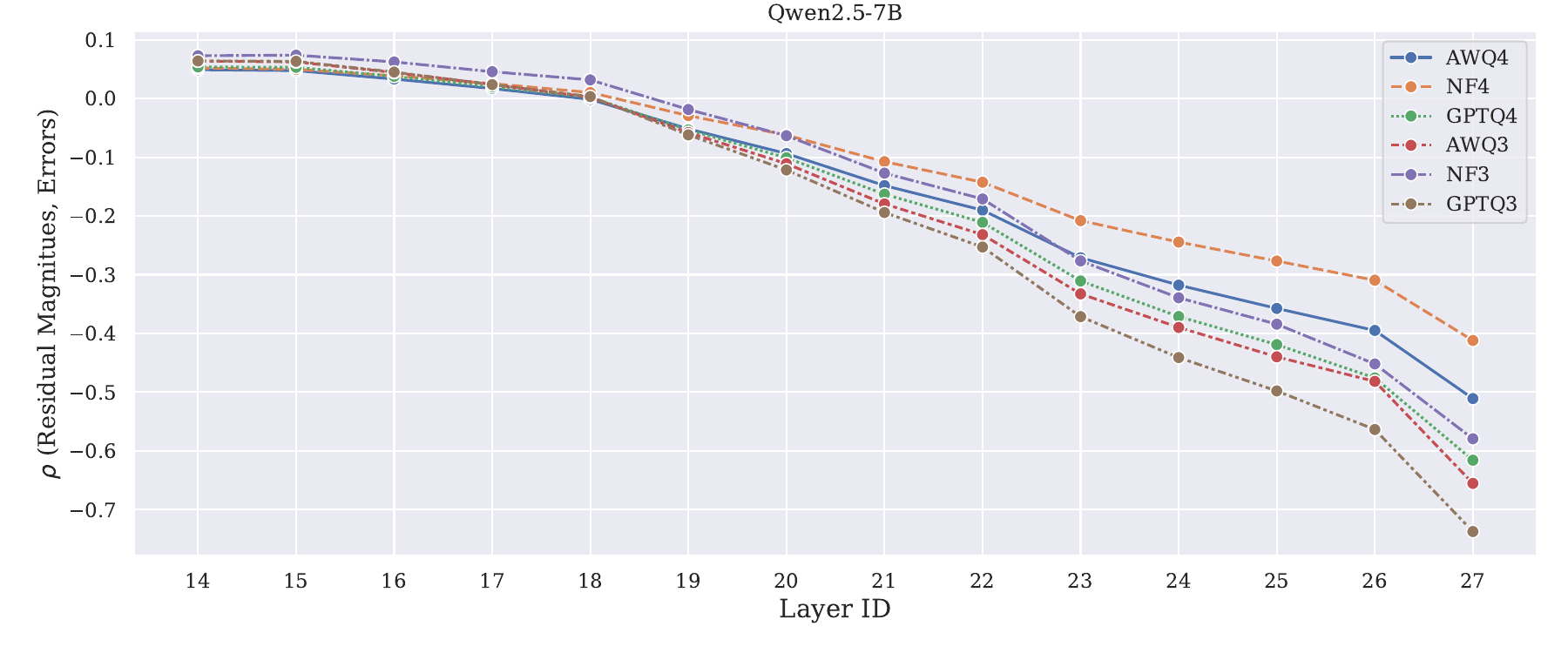}
    \caption{Correlations between the quantization errors $\EDs(\Ds ; \QLLM)$ and the residual magnitudes $\Norm (\Ds; \LLM; l)$ across the upper half layers of Qwen2.5-7B. All methods show negative correlations in the last few layers.}
    \label{fig:app-norm-Qwen2.5-7b}
\end{figure*}

%% file: latex/Figures/app-corr-norms-Meta-Llama-3-8B.tex
\begin{figure*}[t!]
    \centering
    \includegraphics[width=1.\linewidth]{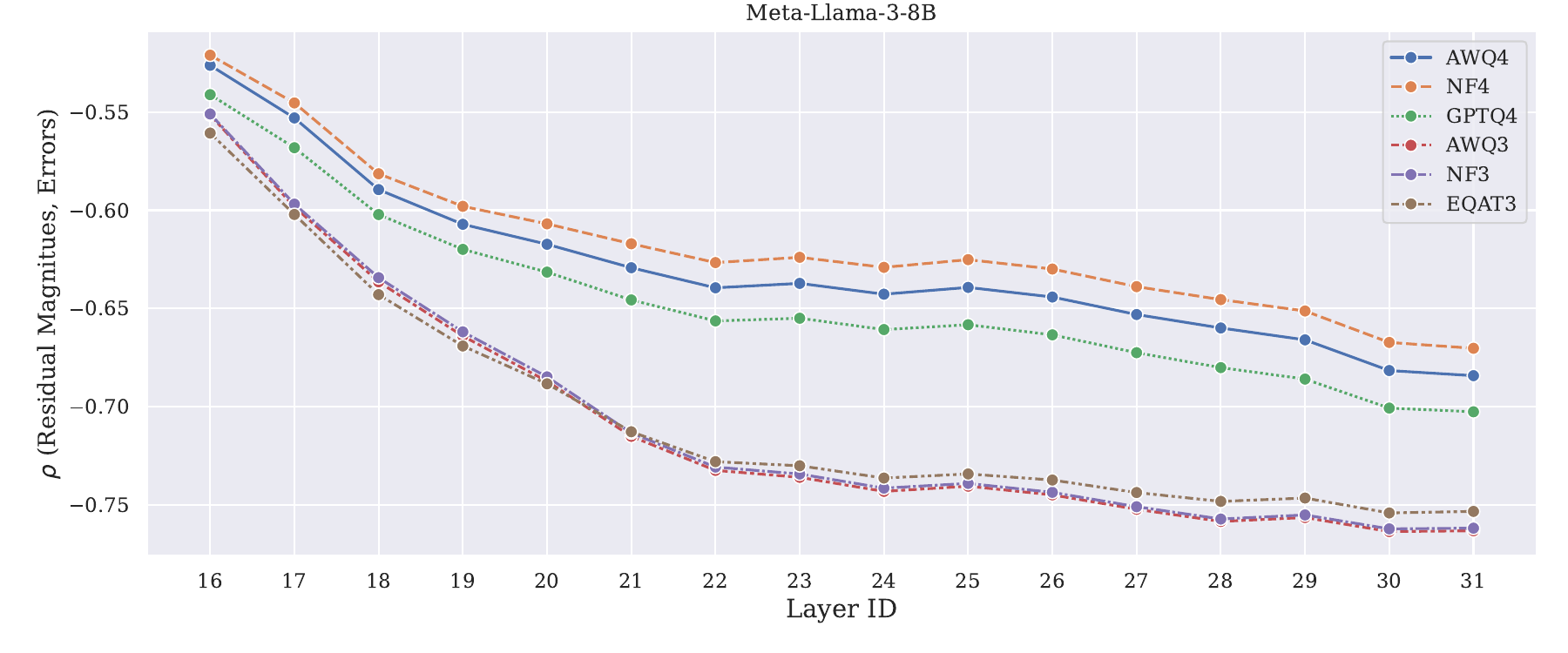}
    \caption{Correlations between the quantization errors $\EDs(\Ds ; \QLLM)$ and the residual magnitudes $\Norm (\Ds; \LLM; l)$ across the upper half layers of Llama3-8B. All methods show negative correlations in the last few layers.}
    \label{fig:app-norm-llama3-8b}
\end{figure*}

%% file: latex/Figures/app-corr-norms-Mistral-Nemo-Base-2407.tex
\begin{figure*}[t!]
    \centering
    \includegraphics[width=1.\linewidth]{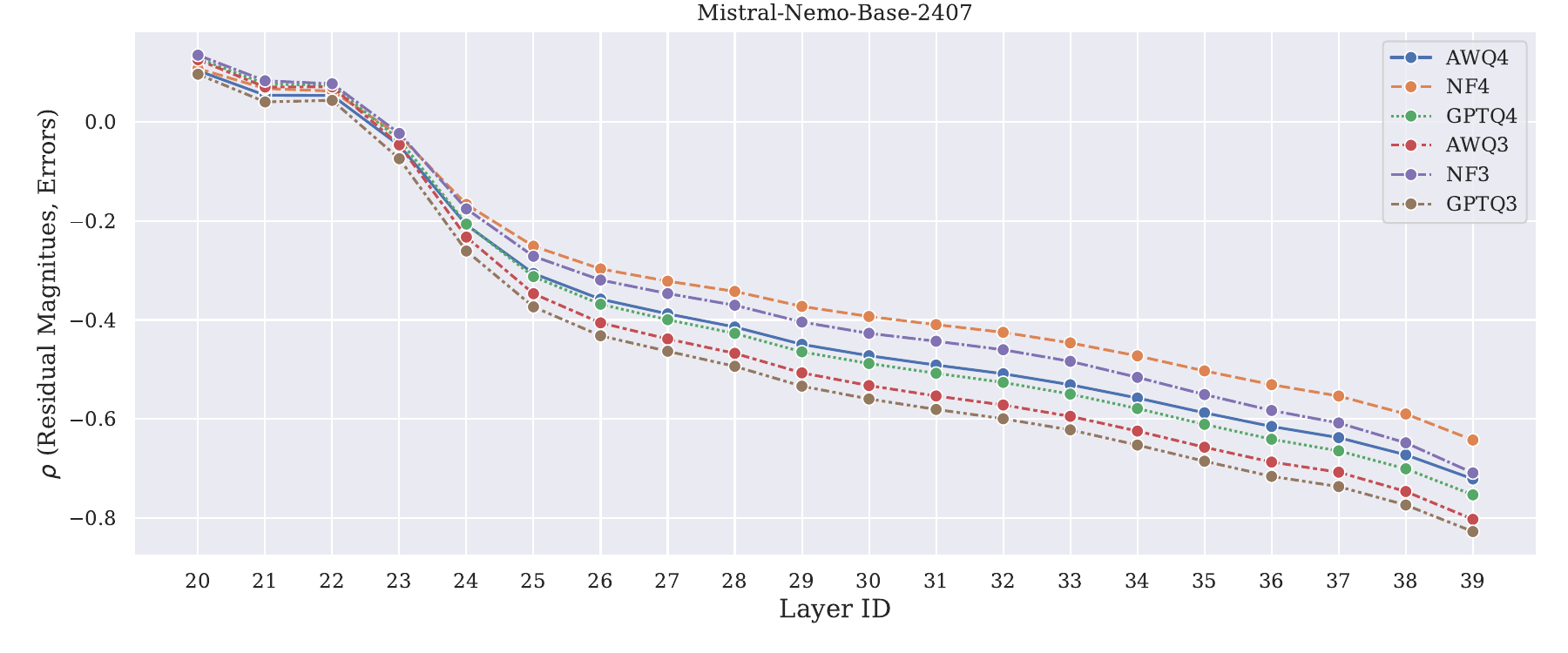}
    \caption{Correlations between the quantization errors $\EDs(\Ds ; \QLLM)$ and the residual magnitudes $\Norm (\Ds; \LLM; l)$ across the upper half layers of Mistral-12B. All methods show negative correlations in the last few layers.}
    \label{fig:app-norm-mistral-12b}
\end{figure*}

%% file: latex/Figures/app-corr-norms-Meta-Llama-3-70B.tex
\begin{figure*}[t!]
    \centering
    \includegraphics[width=1.\linewidth]{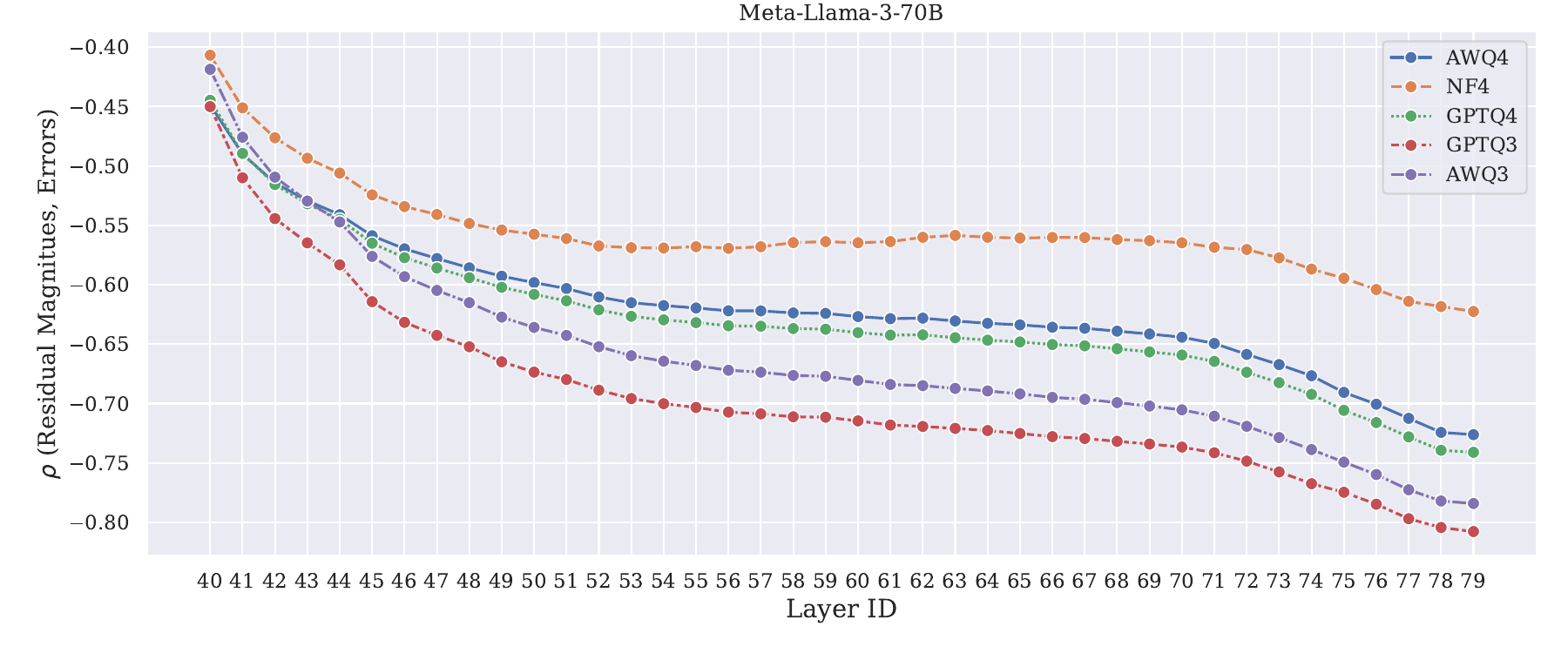}
    \caption{Correlations between the quantization errors $\EDs(\Ds ; \QLLM)$ and the residual magnitudes $\Norm (\Ds; \LLM; l)$ across the upper half layers of Llama3-70B. All methods show negative correlations in the last few layers.}
    \label{fig:app-norm-llama3-70b}
\end{figure*}

%% file: latex/Figures/app-ee-Qwen2.5-7B.tex
\begin{figure*}[t!]
    \centering
    \includegraphics[width=1.\linewidth]{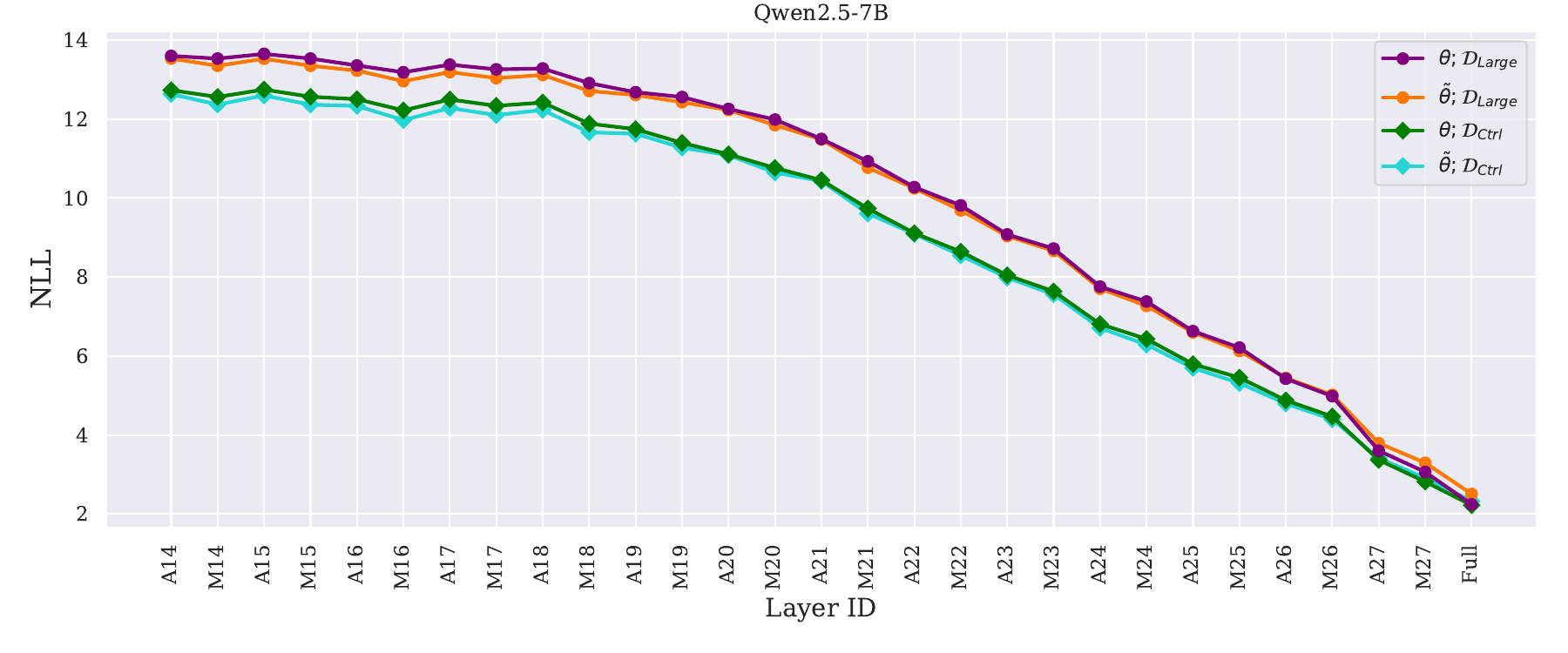}
    \caption{Early Exiting results on Llama3-8B. We decode the $\MLP$ (A\#) and $\Simple$ (M\#) across layers.}
    \label{fig:app-ee-qwen}
\end{figure*}

%% file: latex/Figures/app-ee-Meta-Llama-3-8B.tex
\begin{figure*}[t!]
    \centering
    \includegraphics[width=1.\linewidth]{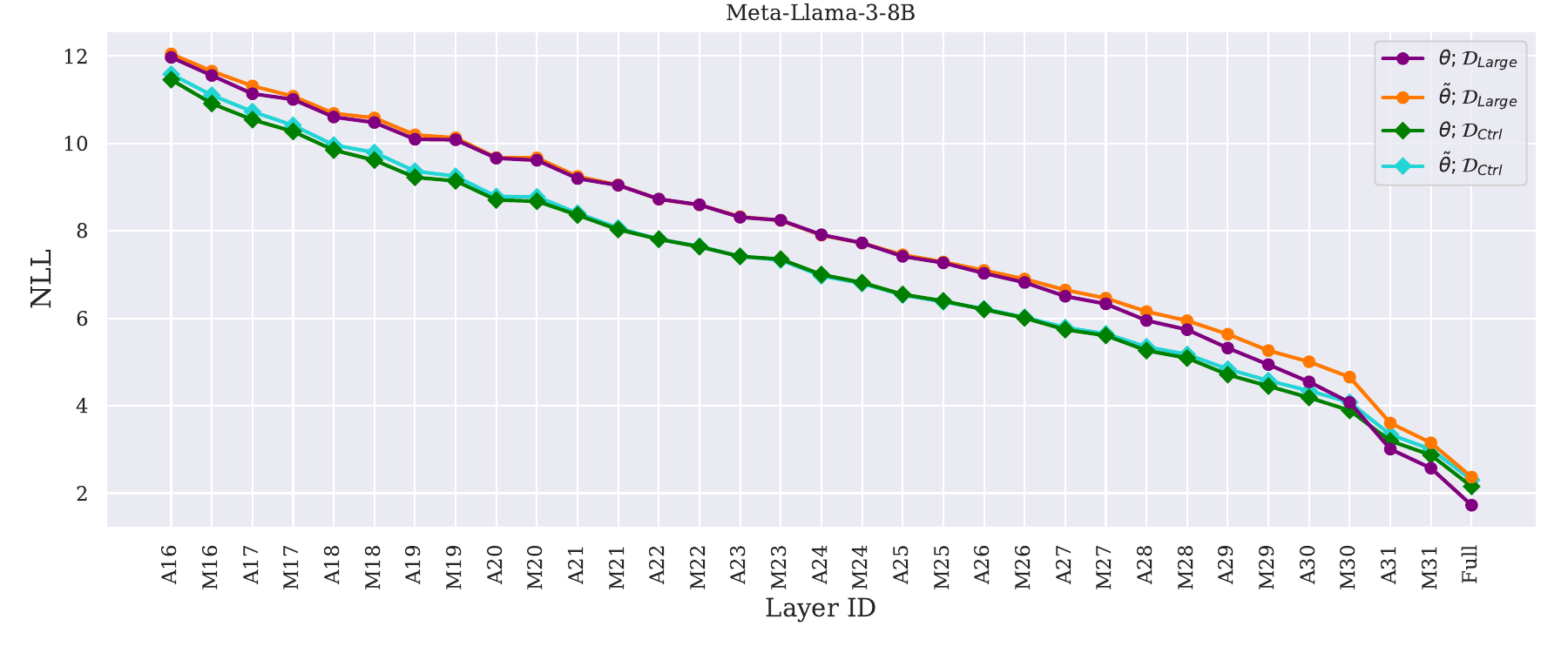}
    \caption{Early Exiting results on Llama3-8B. We decode the inputs to $\MLP$ (A\#) and $\Simple$ (M\#) across layers.}
    \label{fig:app-ee-llama3-8b}
\end{figure*}

%% file: latex/Figures/app-ee-Mistral-Nemo-Base-2407.tex
\begin{figure*}[t!]
    \centering
    \includegraphics[width=1.\linewidth]{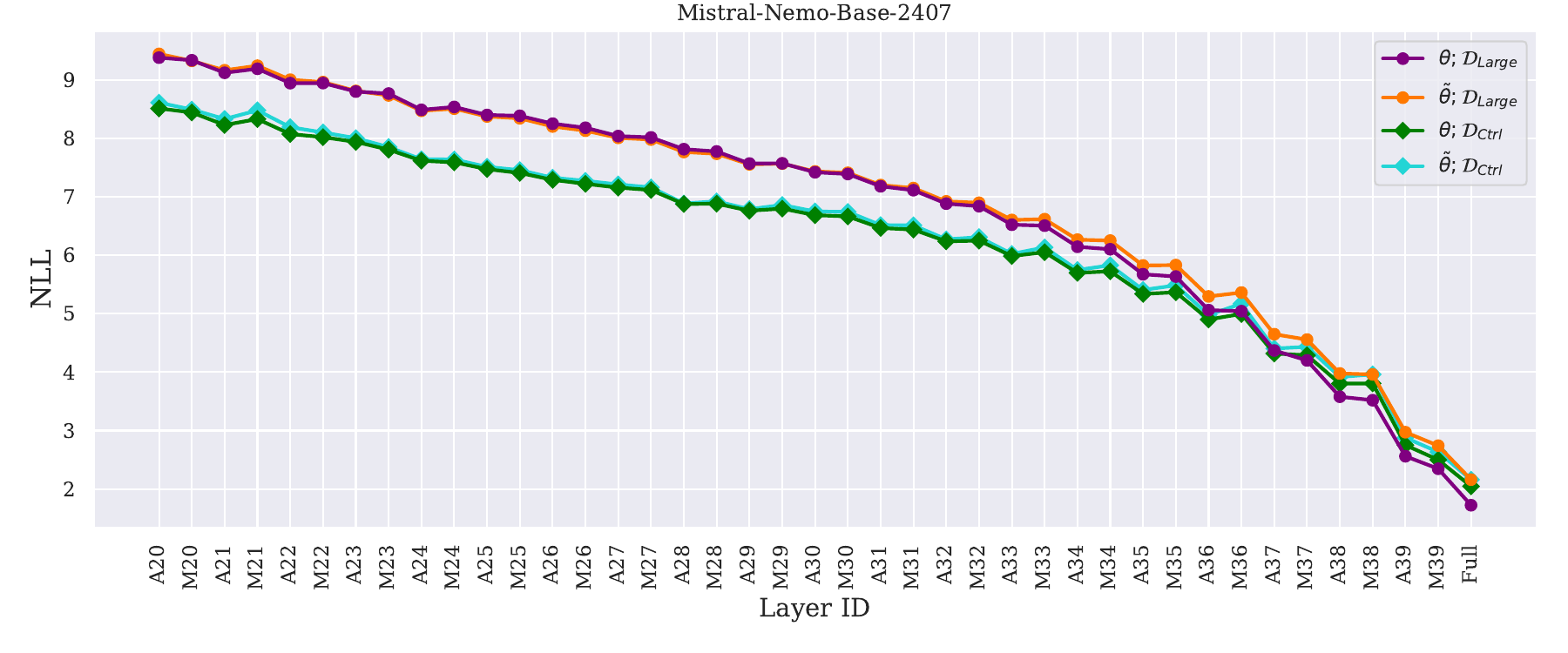}
    \caption{Early Exiting results on Llama3-8B. We decode the $\MLP$ (A\#) and $\Simple$ (M\#) across layers.}
    \label{fig:app-ee-mistral}
\end{figure*}

%% file: latex/Figures/app-ee-Meta-Llama-3-70B.tex
\begin{figure*}[t!]
    \centering
    \includegraphics[width=1.\linewidth]{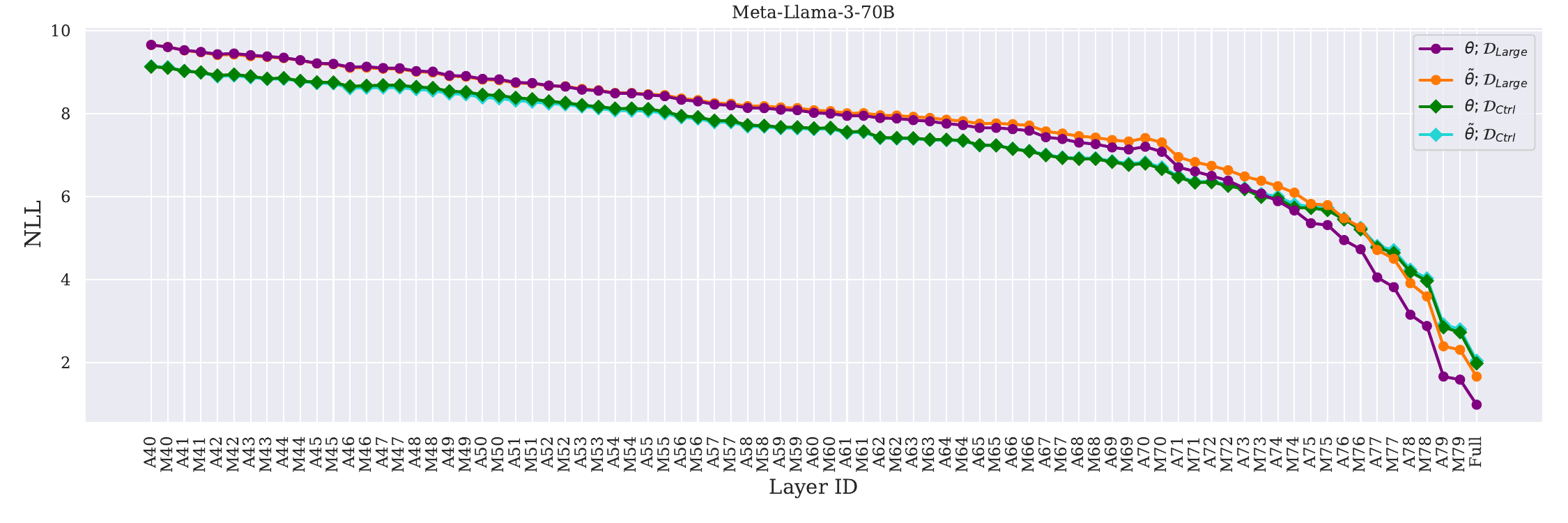}
    \caption{Early Exiting results on Llama3-8B. We decode$\MLP$ (A\#) and $\Simple$ (M\#) across layers.}
    \label{fig:app-ee-llama3-70b}
\end{figure*}

%% file: latex/Figures/app-rare.tex
\begin{figure*}[t!]
    \centering
    \includegraphics[width=1.\linewidth]{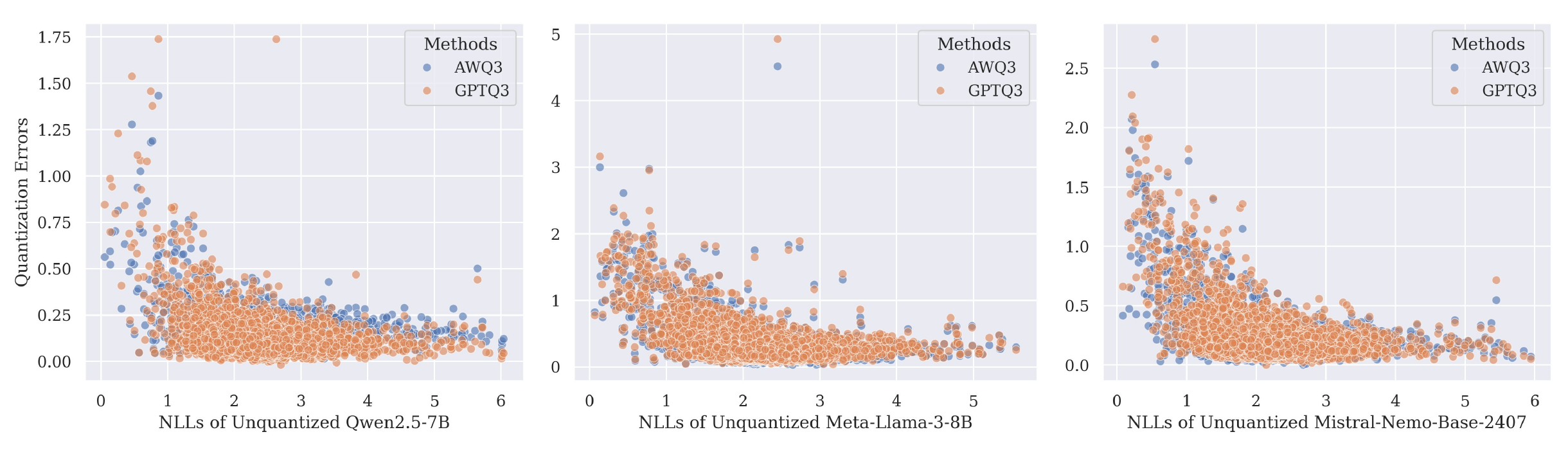}
    \caption{NLLs before quantization vs. quantization errors on different LLMs. Each dot represents a sample from FineWeb. In general, the low-frequency data (x-axis values $\ge 4$) have smaller errors.
    On the other hand, many samples that suffer from large quantization errors have small NLLs before quantization.}
    \label{fig:app-rare}
\end{figure*}

%% file: latex/Figures/app-popqa-delta.tex
\begin{figure*}[t!]
    \centering
    \includegraphics[width=1.\linewidth]{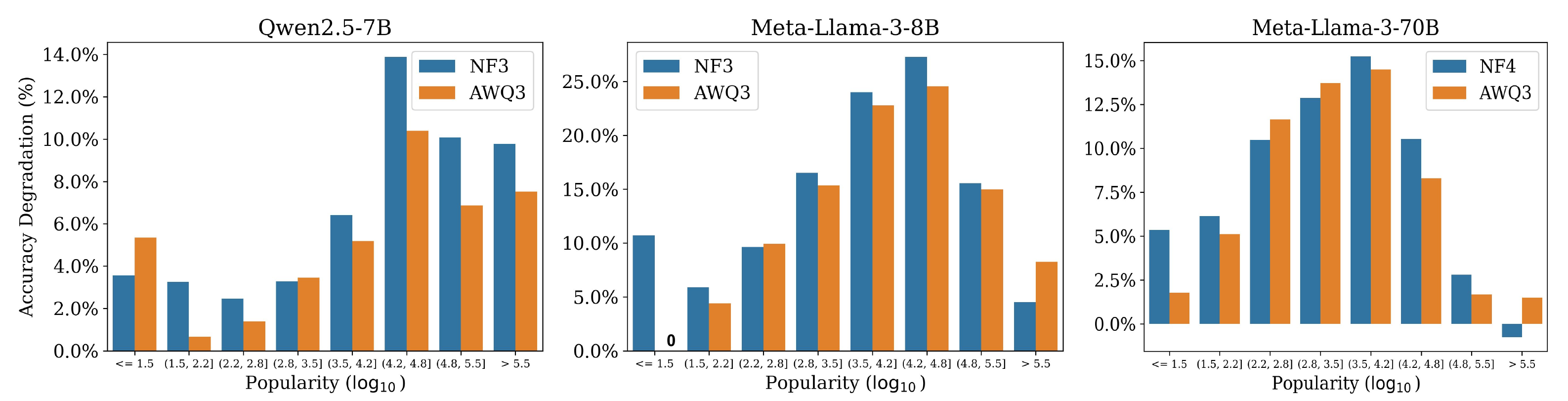}
    \caption{Accuracy degradation of different models on PopQA.}
    \label{fig:app-popqa-delta}
\end{figure*}

%% file: latex/Figures/app-popqa-abs.tex
\begin{figure*}[t!]
    \centering
    \includegraphics[width=1.\linewidth]{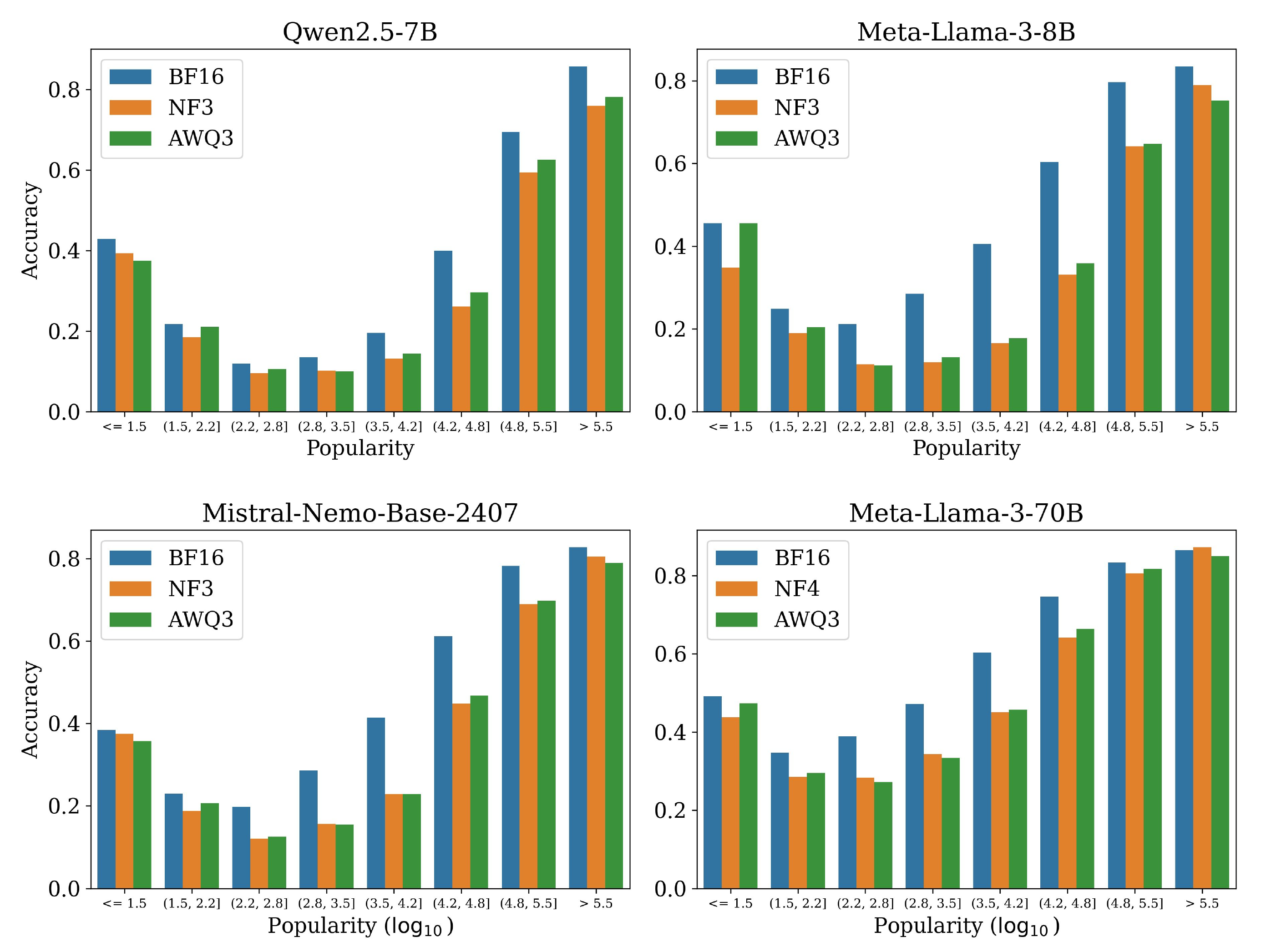}
    \caption{Absolute accuracies of different models on PopQA.}
    \label{fig:app-popqa-abs}
\end{figure*}

%% file: latex/Figures/app-popqa-data.tex
\begin{figure*}[t!]
    
     \begin{minipage}[t]{0.48\textwidth}\includegraphics[width=\linewidth]{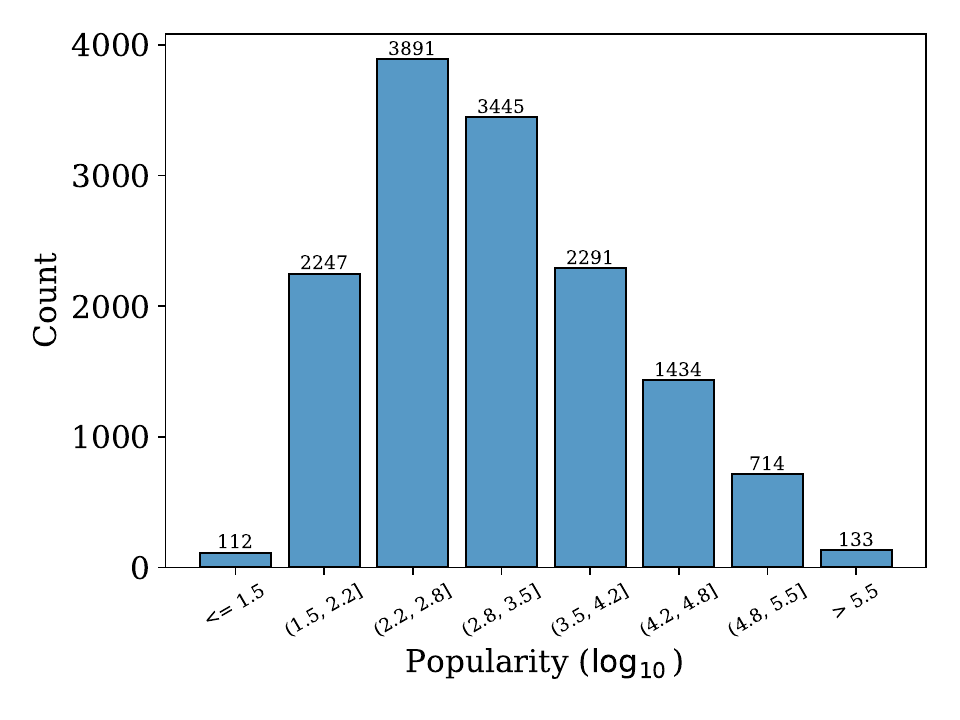}
     \centering
    \caption{The histogram of the PopQA dataset. We partition the dataset into 8 buckets based on the sample popularity and ensure that each partition contains at least 100 samples.}
    \label{fig:app-popqa-data}
    \end{minipage}
\end{figure*}

%% file: latex/Tables/app_gamma_stats.tex
\begin{table*}[!t]
\begin{center}
\begin{small}
\begin{NiceTabular}{lclcc}
    \CodeBefore
    \rectanglecolor{bgpurple}{1-1}{61-2}
    \rectanglecolor{bgyellow}{1-3}{61-3}
    \rectanglecolor{bggreen}{1-4}{61-4}
    \rectanglecolor{bgblue}{1-5}{61-6}
    \Body
\toprule
& \textbf{Layer} & \textbf{Rank of Outliers'} \boldmath{$|\gamma|$} &    \textbf{Median} \boldmath{$|\gamma|$}   & \textbf{Outliers'} \boldmath{$|\gamma|$} \\
\midrule
\multirow{40}{*}{\rotatebox[origin=c]{90}{\textbf{Llama3-70B}}}
& 40 & [2, 9, 1, 10, 13] & 0.1553 & [0.0195, 0.0376, 0.0186, 0.0386, 0.0413]\\
& 41 & [2, 8, 1, 9, 10] & 0.1582 & [0.0170, 0.0312, 0.0142, 0.0320, 0.0330]\\
& 42 & [5, 1, 7, 10, 8] & 0.1602 & [0.0215, 0.0128, 0.0271, 0.0312, 0.0292]\\
& 43 & [1, 6, 2, 8, 9] & 0.1621 & [0.0117, 0.0275, 0.0154, 0.0315, 0.0317]\\
& 44 & [1, 7, 2, 9, 8] & 0.1631 & [0.0171, 0.0369, 0.0209, 0.0396, 0.0374]\\
& 45 & [1, 7, 10, 2, 7] & 0.1660 & [0.0130, 0.0383, 0.0420, 0.0209, 0.0383]\\
& 46 & [1, 7, 14, 12, 2] & 0.1689 & [0.0088, 0.0342, 0.0405, 0.0396, 0.0223]\\
& 47 & [1, 7, 8, 3, 14] & 0.1699 & [0.0090, 0.0320, 0.0364, 0.0206, 0.0396]\\
& 48 & [1, 6, 8, 2, 9] & 0.1709 & [0.0098, 0.0359, 0.0420, 0.0247, 0.0427]\\
& 49 & [1, 5, 12, 3, 12] & 0.1738 & [0.0128, 0.0354, 0.0476, 0.0297, 0.0476]\\
& 50 & [1, 6, 13, 3, 8] & 0.1758 & [0.0100, 0.0327, 0.0437, 0.0273, 0.0369]\\
& 51 & [1, 6, 11, 3, 8] & 0.1777 & [0.0117, 0.0369, 0.0471, 0.0312, 0.0427]\\
& 52 & [1, 7, 8, 3, 19] & 0.1787 & [0.0126, 0.0422, 0.0454, 0.0315, 0.0591]\\
& 53 & [1, 6, 11, 3, 14] & 0.1816 & [0.0136, 0.0337, 0.0422, 0.0255, 0.0461]\\
& 54 & [1, 6, 9, 2, 11] & 0.1826 & [0.0138, 0.0293, 0.0376, 0.0248, 0.0388]\\
& 55 & [1, 5, 9, 3, 11] & 0.1846 & [0.0110, 0.0330, 0.0427, 0.0277, 0.0454]\\
& 56 & [1, 6, 9, 3, 13] & 0.1865 & [0.0138, 0.0378, 0.0439, 0.0297, 0.0471]\\
& 57 & [1, 6, 9, 15, 3] & 0.1885 & [0.0124, 0.0344, 0.0408, 0.0459, 0.0282]\\
& 58 & [1, 6, 9, 15, 14] & 0.1904 & [0.0121, 0.0322, 0.0388, 0.0457, 0.0437]\\
& 59 & [1, 5, 12, 23, 14] & 0.1934 & [0.0206, 0.0420, 0.0559, 0.0684, 0.0583]\\
& 60 & [1, 6, 12, 27, 14] & 0.1953 & [0.0197, 0.0461, 0.0547, 0.0776, 0.0554]\\
& 61 & [1, 7, 12, 24, 14] & 0.1982 & [0.0177, 0.0459, 0.0532, 0.0693, 0.0562]\\
& 62 & [1, 6, 22, 14, 20] & 0.2002 & [0.0160, 0.0435, 0.0659, 0.0579, 0.0635]\\
& 63 & [1, 6, 12, 22, 15] & 0.2031 & [0.0199, 0.0530, 0.0640, 0.0781, 0.0679]\\
& 64 & [1, 8, 26, 18, 16] & 0.2070 & [0.0272, 0.0608, 0.0903, 0.0757, 0.0723]\\
& 65 & [1, 2, 9, 8, 26] & 0.2109 & [0.0225, 0.0403, 0.0635, 0.0603, 0.0903]\\
& 66 & [1, 9, 5, 6, 22] & 0.2119 & [0.0259, 0.0591, 0.0496, 0.0520, 0.0825]\\
& 67 & [1, 10, 6, 26, 7] & 0.2158 & [0.0266, 0.0625, 0.0537, 0.0918, 0.0559]\\
& 68 & [1, 9, 5, 27, 6] & 0.2227 & [0.0293, 0.0762, 0.0659, 0.1079, 0.0674]\\
& 69 & [1, 9, 2, 36, 7] & 0.2266 & [0.0294, 0.0732, 0.0425, 0.1201, 0.0688]\\
& 70 & [1, 12, 28, 6, 7] & 0.2285 & [0.0311, 0.0737, 0.1167, 0.0654, 0.0669]\\
& 71 & [1, 28, 16, 6, 7] & 0.2344 & [0.0312, 0.1328, 0.1006, 0.0771, 0.0776]\\
& 72 & [1, 33, 11, 9, 3] & 0.2441 & [0.0430, 0.1650, 0.0947, 0.0918, 0.0649]\\
& 73 & [1, 36, 9, 23, 3] & 0.2520 & [0.0483, 0.1729, 0.1089, 0.1455, 0.0723]\\
& 74 & [1, 36, 8, 18, 3] & 0.2598 & [0.0459, 0.1846, 0.1060, 0.1367, 0.0767]\\
& 75 & [1, 44, 11, 10, 3] & 0.2676 & [0.0796, 0.2139, 0.1250, 0.1230, 0.0923]\\
& 76 & [1, 43, 3, 15, 15] & 0.3047 & [0.0486, 0.2217, 0.0981, 0.1533, 0.1533]\\
& 77 & [1, 2, 52, 4, 5] & 0.2891 & [0.0554, 0.0820, 0.2080, 0.0869, 0.0913]\\
& 78 & [1, 4, 2, 20, 6] & 0.2812 & [0.0781, 0.0972, 0.0923, 0.1562, 0.1138]\\
& 79 & [9, 4, 1, 3, 13] & 0.2236 & [0.0688, 0.0500, 0.0309, 0.0454, 0.0747]\\

\midrule
\multirow{20}{*}{\rotatebox[origin=c]{90}{\textbf{Mistral-12B}}}
& 20 & [1, 6, 22, 376, 4000] & 0.8594 & [0.001640, 0.137695, 0.330078, 0.648438, 0.902344]\\
& 21 & [1, 5, 22, 289, 2463] & 0.9062 & [0.001648, 0.113770, 0.324219, 0.644531, 0.906250]\\
& 22 & [1, 2, 22, 283, 2402] & 0.9453 & [0.002487, 0.044434, 0.369141, 0.687500, 0.941406]\\
& 23 & [1, 2, 22, 308, 3740] & 0.9570 & [0.001511, 0.089355, 0.410156, 0.726562, 0.988281]\\
& 24 & [2, 1, 261, 21, 4599] & 1.0078 & [0.080078, 0.001457, 0.730469, 0.365234, 1.062500]\\
& 25 & [2, 1, 167, 20, 1615] & 1.0469 & [0.102539, 0.017334, 0.691406, 0.359375, 1.023438]\\
& 26 & [2, 1, 184, 21, 1145] & 1.0859 & [0.092285, 0.001183, 0.722656, 0.390625, 1.039062]\\
& 27 & [5, 1, 151, 21, 840] & 1.1172 & [0.140625, 0.025146, 0.722656, 0.410156, 1.046875]\\
& 28 & [2, 1, 150, 22, 792] & 1.1484 & [0.094727, 0.005585, 0.757812, 0.429688, 1.070312]\\
& 29 & [2, 1, 175, 21, 945] & 1.1875 & [0.121582, 0.016113, 0.785156, 0.453125, 1.132812]\\
& 30 & [2, 1, 181, 19, 1080] & 1.2266 & [0.116211, 0.009583, 0.855469, 0.458984, 1.187500]\\
& 31 & [2, 1, 211, 3486, 21] & 1.2578 & [0.129883, 0.000584, 0.933594, 1.273438, 0.488281]\\
& 32 & [2, 1, 212, 4982, 18] & 1.2812 & [0.153320, 0.000881, 1.000000, 1.335938, 0.458984]\\
& 33 & [2, 1, 200,  \textcolor{orange}{5098}, 18] & 1.2812 & [0.154297, 0.001419, 1.054688, 1.421875, 0.542969]\\
& 34 & [1, 2,  \textcolor{orange}{5111}, 666, 21] & 1.3125 & [0.000793, 0.210938, 1.640625, 1.273438, 0.636719]\\
& 35 & [1, 2,  \textcolor{orange}{5112, 5054, 5110}] & 1.3594 & [0.110352, 0.213867, 1.773438, 1.484375, 1.742188]\\
& 36 & [1, 2,  \textcolor{orange}{5114, 5111, 5064}] & 1.3906 & [0.001472, 0.277344, 1.984375, 1.906250, 1.578125]\\
& 37 & [1, 19, \textcolor{orange}{5116, 5107, 5095}] & 1.3828 & [0.001205, 0.808594, 2.328125, 2.109375, 1.867188]\\
& 38 & [1, \textcolor{orange}{5119, 5115, 5118, 5103}] & 1.4219 & [0.316406, 3.078125, 2.921875, 3.031250, 2.359375]\\
& 39 & [1, 28, 61, 40,  \textcolor{orange}{5120}] & 1.6172 & [0.601562, 1.265625, 1.476562, 1.406250, 5.437500]\\
\bottomrule
\end{NiceTabular}
\end{small}
\end{center}
\caption{We rank the RMSNorm weights $\gamma$ in descending order by the absolute value of each entry. We then compare the median weight with those of the outlier dimensions. We list the ranks and weight values corresponding to the 1st through 5th severest outliers. In Llama3-70B, all outlier dimensions have weights far below the median, suggesting that the model uses RMSNorm weights to suppress outliers.
In Mistral-12B, the severest outliers have small weights (rank 1 or 2) across layers, but the other outliers sometimes have large weights (rank \textcolor{orange}{$>5000$}).}
\label{table:app_gamma_stats}
\end{table*}